\documentclass[final]{elsarticle}

\usepackage{lineno,hyperref}
\modulolinenumbers[5]

\journal{Information Sciences}

\usepackage{amsthm}
\usepackage{pdfpages}
\usepackage{stmaryrd}
\usepackage{enumitem}
\usepackage{comment}
\usepackage{times}

\usepackage[ruled,vlined,linesnumbered]{algorithm2e}

\usepackage{graphicx,booktabs,listings,caption,subcaption,tikz,amsmath,amssymb,placeins,subfiles,tabto,scalerel,stackengine}
\usepackage{xcolor,colortbl}
\usepackage{multirow,float}

\usepackage{mathtools,centernot}
\usepackage{tikz, trimclip}

\usepackage{makecell}
\usepackage{cleveref}
\usepackage{pifont}

\newtheorem{definition}{Definition}

\definecolor{azure}{rgb}{0.0, 0.5, 1.0}
\definecolor{cadmiumorange}{rgb}{0.93, 0.53, 0.18}
\definecolor{cadmiumgreen}{rgb}{0.0, 0.42, 0.24}
\definecolor{cadmiumred}{rgb}{0.89, 0.0, 0.13}
\definecolor{cadmiumyellow}{rgb}{1.0, 0.96, 0.0}
\definecolor{applegreen}{rgb}{0.55, 0.71, 0.0}

\definecolor{otr}{rgb}{0.9725490196078431, 0.807843137254902, 0.8}

\definecolor{r}{rgb}{0.9568627450980393, 0.3686274509803922, 0.34509803921568627}
\definecolor{g}{rgb}{0.0, 0.5803921568627451, 0.2980392156862745}
\definecolor{p}{rgb}{0.5764705882352941, 0.47058823529411764, 0.8901960784313725}
\definecolor{y}{rgb}{0.9529411764705882, 0.7294117647058823, 0.2196078431372549}
\definecolor{o}{rgb}{0.9568627450980393, 0.4980392156862745, 0.2980392156862745}
\definecolor{pi}{rgb}{0.9803921568627451, 0.4980392156862745, 0.6941176470588235}

\newcolumntype{L}{>{$}c<{$}}
\newcommand{\mode}       {\mu}
\newcommand{\F}          {\mathcal{F}}

\newcommand{\Lang}       {\mathcal{L}}
\newcommand{\M}          {\mathcal{M}}

\newcommand{\R}          {\mathcal{R}}
\newcommand{\Ob}         {\mathcal{O}}
\newcommand{\V}          {\mathcal{V}}
\newcommand{\Nat}        {\mathbb{N}}

\newcommand{\supp}       {\mathit{supp}}

\newcommand{\nunet}      {$\nu$-net}
\newcommand{\nunets}     {$\nu$-nets}

\newcommand{\al}         {\gamma}  
\newcommand{\relal}      {\tilde\gamma} 
\newcommand{\run}        {\varphi}
\newcommand{\relL}      {\tilde{L}} 
\newcommand{\relrun}      {\tilde{\run}} 
\newcommand{\Var}          {\mathit{Var}}

\newcommand{\Lissue}[2] {$\text{I}_\text{#1}^\text{#2}$}

\newcommand{\proj}              {{\upharpoonright}}

\DeclareMathOperator{\obj}{objects}

\DeclareMathOperator{\role}{role}

\DeclareMathOperator{\match}{match}

\newcommand{\set}[1]{\{ #1 \}}
\newcommand{\sset}[1]{\left\{ #1 \right\}}
\newcommand{\mset}[1]{[ #1 ]}

\newcommand{\seq}[1]{\langle #1 \rangle}

\newcommand{\stup}[1]{\left(#1 \right)}

\newcommand{\smid}{\,\middle|\,}

\newcommand{\pre}[1]             {\prescript{\bullet}{}{#1}}
\newcommand{\post}[1]            {#1^\bullet}

\newcommand{\Ie}{I.e., }
\newcommand{\ie}{i.e., }
\newcommand{\Eg}{E.g., }
\newcommand{\eg}{e.g., }

\newcommand{\cf}{c.f., }

\newcommand{\id}                {\mathit{id}}

\newcommand{\orderhome}[1]{\substack{\text{order}\\\text{home}}^{#1}}
\newcommand{\orderdepot}[1]{\substack{\text{order}\\\text{depot}}^{#1}}
\newcommand{\ring}[1]{\substack{\text{ring}}^{#1}}
\newcommand{\registerdepot}[1]{\substack{\text{register}\\\text{depot}}^{#1}}
\newcommand{\deliverhome}[1]{\substack{\text{deliver}\\\text{home}}^{#1}}
\newcommand{\deliverdepot}[1]{\substack{\text{deliver}\\\text{depot}}^{#1}}
\newcommand{\collect}[1]{\substack{\text{collect}}^{#1}}
\newcommand{\tauone}[1]{\substack{\tau_1}^{#1}}
\newcommand{\tautwo}[1]{\substack{\tau_2}^{#1}}
\newcommand{\tstart}[1]{\substack{\text{start}}^{#1}}
\newcommand{\tstop}[1]{\substack{\text{stop}}^{#1}}
\newcommand{\create}[1]{\substack{\text{create}}^{#1}}
\newcommand{\destroy}[1]{\substack{\text{destroy}}^{#1}}

\newcommand{\tauc}{\tau_\text{create}}

\begin{document}

\begin{frontmatter}

\title{In System Alignments we Trust!\\
Explainable Alignments via Projections}

\author{Dominique Sommers}
\author{Natalia Sidorova}
\author{Boudewijn van Dongen}

\address{Eindhoven University of Technology, Mathematics and Computer Science, Eindhoven, the Netherlands\\\{\href{mailto:d.sommers@tue.nl}{d.sommers}, \href{mailto:n.sidorova@tue.nl}{n.sidorova}, \href{mailto:b.f.v.dongen@tue.nl}{b.f.v.dongen}\}@tue.nl}






\begin{abstract}
Alignments are a well-known process mining technique for reconciling system logs and normative process models. Evidence of certain behaviors in a real system may only be present in one representation -- either a log or a model -- but not in the other.
Since for processes in which multiple entities, like objects and resources, are involved in the activities, their interactions affect the behavior and are therefore essential to take into account in the alignments.

Additionally, both logged and modeled representations of reality may be imprecise and only partially represent some of these entities, but not all. In this paper, we introduce the concept of ``relaxations'' through projections for alignments to deal with partially correct models and logs. Relaxed alignments help to distinguish between trustworthy and untrustworthy content of the two representations (the log and the model) to achieve a better understanding of the underlying process and expose quality issues.
\end{abstract}
\begin{keyword}
Petri nets, Conformance checking, Interacting objects, Projections.
\end{keyword}

\end{frontmatter}

\section{Introduction}\label{sec:introduction}
Conformance checking is a branch of process mining focusing on understanding and improving processes by comparing the behavior of the process recorded in an event log with a normative process model representing the expected behavior. Both the recorded and modeled behavior may contain imperfections and even contradicting information regarding the system's true nature, which is further complicated by the presence of multiple entities simultaneously involved in real-life processes and interacting with each other in complex ways~\cite{van2020discovering,fahland2019describing,van2003soundness,sommers2022aligning}.

\begin{figure}[tb]
\centering
\includegraphics[scale=0.6]{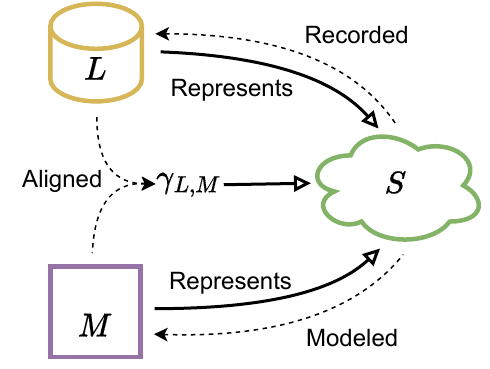}
    \caption{System log $L$ and process model $M$ in relation to the system $S$.}
    \label{fig:SLM}
\end{figure}

In process mining, the process is generally an unknown entity with complex behavior manifested by its activities, which occur at different times and are executed by various resources on different objects. This behavior includes decision points, concurrency, and long-term dependencies between activities and/or resources. Logging mechanisms reveal some aspects of this behavior, while normative process models represent the expected system behavior, as illustrated in Figure~\ref{fig:SLM}. The log representation of the system tends to be more concrete than a normative process model. For example, the former may indicate the concrete resources that performed a specific activity, while a model may only specify the roles of the required resources; the log may show that activity $a$ was executed before activity $b$, while the model would indicate that they are parts of parallel branches.

Logs and/or models are often presumed to be trustworthy representations of a system and used in process analysis under this premise. For instance, in model repair, logs are entirely trusted, and log information is utilized to modify a process model and make it more aligned with the observed behavior~\cite{fahland2015model}. Similarly, in log repair, data quality issues in a log are identified based on a trusted model~\cite{rogge2014temporal,rogge2013improving,wang2015cleaning}. However, both log and model representations are potentially imprecise~\cite{bose2013wanna,carmona2018conformance,wynn2019responsible}, and it is typically unjustifiable to trust that one of them accurately represents the system. To address this issue, Rogge-Solti et al.~\cite{rogge2016log} introduced the notion of generalized conformance checking, which combines the comparison of observed and modeled behavior with log and model repair. This approach aims to balance trust in the two representations by iteratively trusting one representation to partially repair the other and vice versa, lifting the assumption of trustworthiness.

In this paper, we demonstrate how aligning a complete system log $L$ and a process model $M$ can improve the representation of reality in the context of processes with multiple perspectives, while taking into consideration potential trust issues with both $L$ and $M$. 
We show how relaxed alignments loosen constraints regarding certain perspectives to accommodate partially compliant behaviors and ultimately to pinpoint sources of deviations more precisely. 
This approach allows us to assess whether the content of $L$ and/or $M$ is likely to be part of the behavior of system $S$. The relaxed alignment distinguishes between the trustworthy and untrustworthy content of $L$ and $M$, providing a better understanding of the underlying process, which can also be further used in subsequent analysis. In previous work~\cite{sommers2024conformance}, we analyzed and proved essential properties of these alignments on which we build in this paper.

We illustrate our approach on a running example, further elaborated in Sec.~\ref{sec:representations}. We consider a package delivery process that can be viewed from multiple perspectives, such as those of the packages, the deliverer, or the warehouse. Our example process consists of the ordering, delivery (at home or at a depot), and collection of packages. Both the process model and the log are imprecise representations of the actual delivery process, which becomes apparent in relaxed alignments.

The paper is organized as follows. In Sec.~\ref{sec:representations}, we describe recorded and modeled behavior as representations of reality. Sec.~\ref{sec:alignments_repr} introduces how alignments provide a richer representation of $S$ by comparing where $L$ and $M$ agree and disagree. In Sec.~\ref{sec:relaxations}, we present the concept of relaxed alignments capable of exposing partial compliance of $L$ and $M$. The implications of our work are discussed in Sec.~\ref{sec:conclusion}.

\section{Preliminaries}\label{sec:preliminaries}
In this section, we start with definitions and notation regarding multisets and partially ordered sets, after which we introduce basic concepts related to Petri nets. Note we present most definitions in this section as reiterations from previous work~\cite{sommers2024conformance} for the sake of self-containedness of this paper.

\begin{definition}{(Multiset)}
A \emph{multiset} $m$ over a set $X$ is $m: X \rightarrow \Nat$. $X^\oplus$ denotes the set of all multisets over $X$. We define the support $\supp(m)$ of a multiset $m$ as the set $\{x \in X \mid m(x) > 0\}$. We list elements of the multiset as $[m(x) \cdot x \mid x \in X]$, and write $|x|$ for $m(x)$, when it is clear from context which multiset it concerns. 

For two multisets $m_1,m_2$ over $X$, we write $m_1 \leq m_2$ if $\forall_{x \in X} m_1(x) \leq m_2(x)$, and $m_1 < m_2$ if $m_1 \leq m_2 \wedge m_1 \neq m_2$. We define $m_1 + m_2 = [(m_1(x) + m_2(x))\cdot x \mid x \in X]$, and $m_1 - m_2 = [\max(0, m_1(x) - m_2(x))\cdot x \mid x \in X]$ for $m_1 \geq m_2$.
\end{definition}

\begin{definition}{(Sequence)}
    A \emph{sequence} $\sigma$ over a set $X$ of length $n \in \Nat$ is $X^n$. With $n > 0$, we write $\sigma = \seq{\sigma_1,\dots,\sigma_n}$, and denote $n$ by $|\sigma|$. $\seq{}$ denotes the empty sequence where $n = 0$. The set of all finite sequences over $X$ is denoted by $X^*$. A projection of a sequence $\sigma$ on a set $Y$ is defined inductively by $\seq{}\proj_Y = \seq{}$ and $(\seq{x}) \cdot \sigma)\proj_Y = \begin{cases} \seq{x}\cdot \sigma\proj_Y & \text{if } x \in Y \\ \sigma\proj_Y & \text{otherwise} \end{cases}$, where $x \in X$ and $\cdot$ is the sequence concatenation operator.
\end{definition}

\begin{definition}{(Partial order, Covering relation, Partially ordered set)}\label{def:poset}
    A \emph{partially ordered set} (\emph{poset}) $X = (\bar X, \prec_X)$ is a pair of a set $\bar X$ and a \emph{partial order} $\prec_X \subseteq X \times X$ (irreflexive, antisymmetric, and transitive).
    We overload the notation and write $x \in X$ if $x \in \bar X$. For $x,y \in X$, we write $x \|_X y$ if $x \nprec y \wedge y \nprec x$ and $x \preceq y$ if $x \prec y \vee x = y$.

    $\pre{x} = \set{y \in \bar X \mid y \prec x}$ and $\post{x} = \set{y \in \bar X \mid x \prec y}$ denote respectively the preset and postset of an element $x \in \bar{X}$. For a subset $\bar{X}' \subseteq \bar{X}$ and an element $x \in \bar{X} \setminus \bar{X}'$, we write $\bar{X}' \prec x$ if for any $y \in \bar{X}'$, $y \prec x$. Similarly for $x \prec \bar{X}'$.
    
    Given a relation $\prec$ that is reflexive, antisymmetric, but not necessarily transitively closed, we define $\prec^+$ to be the smallest transitively closed relation containing $\prec$. Thus $\prec^+$ is a partial order with $\prec \subseteq \prec^+$. The \emph{covering relation} $\lessdot \subset \prec$ is the transitive reduction of partial order $\prec$ which is the smallest subset of $\prec$ with $\lessdot^+ = \prec$, \ie $\lessdot = \sset{ (x, y) \in \prec \smid \forall_{(x,z) \in \prec} z \nprec y }$.

    We define $X^<$ as the set of all sequences with elements from $\bar X$ that respects the partial order, \ie for any $\sigma \in X^<$ we have $|\sigma| = |\bar X|$, $\set{x \in \sigma} = \bar X$ and $\prec_X \subseteq <_\sigma$.
\end{definition}

\begin{definition}{(Set operations, minimum and maximum on posets, poset cuts)}
    We extend the standard set operations of union, intersection, difference and subsets to posets: for any two posets $X$ and $Y$, $X \circ Y = (\bar{X} \circ \bar{Y}, (\prec_X \circ \prec_Y)^+)$, with $\circ \in \set{\cup, \cap, \setminus}$ and $Y \subseteq X$ iff $\bar{Y} \subseteq \bar{X}$ and $\prec_Y = \prec_X \cap (\bar{Y} \times \bar{Y})$.

    Let $X = (\bar X, \prec_X)$ be a partially ordered set. The \emph{minimum} and \emph{maximum} elements of $X$ are defined by $\min(X) = \set{x \mid x \in X, \forall_{y \in X} y \nprec x}$ and $\max(X) = \set{x \mid x \in X, \forall_{y \in X} x \nprec y}$.

    A \emph{cut} in $X$ is a partition $(X_1, X_2)$ of $X$ with $X_1 = (\bar{X}_1, \prec_{X_1})$ and $X_1 = (\bar{X}_2, \prec_{X_2})$ with the following properties: (1) $\bar{X}_1 \uplus \bar{X}_2 = \bar X$, (2) if $x \in \bar{X}_1$ and $y \prec_L x$, then $y \in \bar{X}_1$, and (3) if $x \in \bar{X}_2$ and $x \prec_L y$, then $y \in \bar{X}_2$.

    A sequence of $k$ cuts $(X_1,\dots,X_{k+1})$ in $X$ is again a partition where for each $1 \leq i \leq k$, $\stup{ \bigcup_{1\leq j \leq i} X_j, \bigcup_{i < j \leq k+1} X_j }$ is a cut.
\end{definition}

\begin{definition}{(Poset projection)}
    For a poset $X = (\bar X, \prec_X)$ and a subset $Y \subseteq \bar X$, we define the \emph{projection} $X\proj_Y$ on $X$ as a subposet of $X$ with only elements from $Y$, \ie $X\proj_Y = \left( \bar X \cap Y, \prec_X \cap (Y \times Y) \right)$.
\end{definition}

\begin{figure}[tb]
\centering
\begin{subfigure}{.32\textwidth}
  \centering
  \includegraphics[scale=0.8]{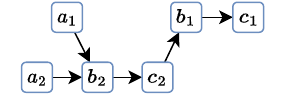}
  \caption{Example poset $A$.}
  \label{fig:prel_poset_A}
\end{subfigure}
\begin{subfigure}{.32\textwidth}
  \centering
  \includegraphics[scale=0.8]{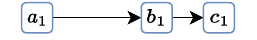}
  \caption{Poset projection $A\proj_X$, with $X = \set{a_1,b_1,c_1}$.}
  \label{fig:poset_A1}
\end{subfigure}
\begin{subfigure}{.32\textwidth}
  \centering
  \includegraphics[scale=0.8]{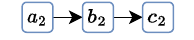}
  \caption{Poset projection $A\proj_Y$ with $Y = \set{a_2,b_2,c_2}$.}
  \label{fig:poset_A2}
\end{subfigure}
\caption{Example of a poset and two projections.}
\label{fig:posets}
\end{figure}
Fig.~\ref{fig:prel_poset_A} depicts a simple poset $A = (\bar{A}, \prec_A)$ with elements $\bar{A} = \set{a_1, b_1,c_1,a_2,b_2,\allowbreak c_2}$ and with partial order $\prec_A = \set{ (a_1,b_1), (b_1,c_1), (a_1,c_1), (a_2,b_2), (b_2,c_2), (a_2,c_2),\allowbreak$ $(a_1, b_2), (a_1, c_2), (a_2, b_1), (a_2, c_1), (b_2, b_1), (b_2, b_1), (c_2, b_1), (c_2, c_1)}$. For clarity reasons, only the covering relation $\lessdot_A = \set{(a_1, b_2), (a_2, b_2), (b_2, c_2), (c_2, b_1), (b_1, c_1)}$ is visualized, as we do as well throughout the rest of this paper. Figs.~\ref{fig:poset_A1} and~\ref{fig:poset_A2} depict example projections of $A$ onto the elements $\set{a_1, b_1, c_1}$ and $\set{a_2, b_2, c_2}$ respectively.

\paragraph{Petri nets} Petri nets can be used to represent, validate, and verify workflow processes to provide insights into how a process behaves~\cite{peterson1981petri}. Process activities are described by labels from alphabet $\Sigma$. Invisible, silent activities are shown with $\tau$-labels. We define $\Sigma^\tau$ as $\Sigma \cup \{\tau\}$.

\begin{definition}{(Labeled Petri net)}\label{def:petri_net}
    A \emph{labeled Petri net}~\cite{murata1989petri} is a tuple $N = (P, T, \F, \ell)$, with sets of places and transitions $P$ and $T$, respectively, such that $P \cap T = \emptyset$, and a multiset of arcs $\F: (P \times T) \cup (T \times P) \rightarrow \Nat$ defining the flow of the net. $\ell: T \rightarrow \Sigma^\tau$  is a \emph{labeling} function, assigning  a label $\ell(t)$ to a transition $t$.

    Given an element $x \in P \cup T$, its \emph{pre- and post-set}, $\pre{x}$ and $\post{x}$, are sets defined by $\pre{x} = \sset{y \in P \cup T \mid \F((y, x)) \geq 1}$ and $\post{x} = \sset{y \in P \cup T \mid \F((x, y)) \geq 1}$ respectively.
\end{definition}

\begin{definition}{(Marking, Enabling and firing of transitions, Reachable markings)}\label{def:marking}
A \emph{marking} $m \in P^\oplus$ of a labeled Petri net $N = (P, T, \F, \ell)$ defines the state of $N$ by assigning a number of tokens to each place.

A transition $t \in T$ is \emph{enabled} for firing in a marking $m$ of net $N$ iff $m \geq \pre{t}$. We denote the \emph{firing} of $t$ in marking $m$  by $m \xrightarrow{t} m'$, where $m'$ is the resulting marking after firing $t$ defined as $m' = m - \pre{t} + \post{t}$. For a transition sequence $\sigma = \seq{t_1,\dots,t_n}$ we write $m \xrightarrow{\sigma} m'$ to denote the consecutive firing of $t_1,\ldots,t_n$. We say that $m'$ is reachable from $m$ and write $m \xrightarrow{*} m'$ if there is some $\sigma \in T^*$ such that $m \xrightarrow{\sigma} m'$.

$\M(N)=P^\oplus$ denotes the set of all markings in net $N$, while $\R(N,m)$ denotes the \emph{set of  markings reachable} in net $N$ from marking $m$.
\end{definition}
\begin{figure}[tb]
    \centering
    \includegraphics[scale=0.6]{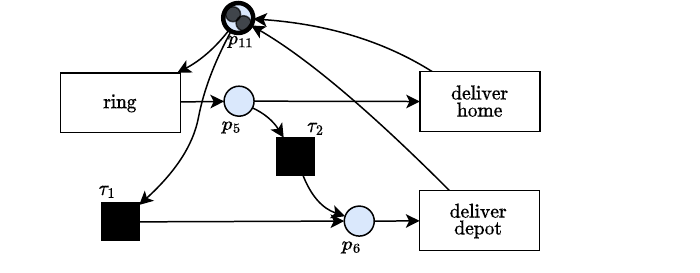}
    \caption{Example labeled Petri net $N_1$ with marking $m$ such that $m(p_5) 
 = m(p_6) = 0$ and $m(p_{11}) = 2$.}
    \label{fig:net}
\end{figure}
Fig.~\ref{fig:net} shows a simple labeled Petri net $N_1 = (P_1, T_1, \F_1, \ell_1)$ modeling the process of a package delivery from the perspective of the deliverer, who delivers packages either at home ($\deliverhome{}$) or at a depot ($\deliverdepot{}$). Prior to a home delivery, the deliverer rings the doorbell ($\ring{}$), and reverts to depot delivery when the client does not answer. In the process, there are two deliverers, depicted by the two tokens in place $p_{11}$. For the running example, we will expand on this process in the following section.

\section{Recorded and modeled behavior as representations of reality}\label{sec:representations}
Figure~\ref{fig:SLM} shows the real process as an unknown entity, denoted as the system $S$. This system generates a system log $L$. A process model $M$ allows for describing and reasoning about process behavior at some abstraction level. In this section, we formalize these entities in order to allow for comparisons to expose where the behaviors of $S$, $L$, and $M$ agree and disagree.

\subsection{Defining reality as \emph{system behavior}}\label{sec:system_behavior}
The core elements of a process' behavior are events. An \emph{event} $e^\id=(a, t, O)^\id$ is an atomic execution of an activity $a \in \Sigma$, from a set of activities $\Sigma$, that occurred at timestamp $t$. We later focus solely on the relative ordering between events, abstracting away from their actual timestamps. $O$ denotes the involved objects and actors of $e$, which is a non-empty multiset $\emptyset < O \leq \Ob$ of a multiset $\Ob$ of object names.

Each object $o \in \Ob$ belongs to exactly one object role $r$ from a set of roles $\R$, given by $\role(o) = r$. We partition multiset $\Ob$ by roles from $\R$ into multisets $\Ob_r$ of object names per object role, \ie $\supp(\Ob) = \biguplus_{r \in \R} \supp(\Ob_r)$. We extend the notation of $\Ob_r$ to $\Ob_R = \mset{\Ob(o) \cdot o \mid o \in \supp(\Ob), \role(o) \in R}$.
Furthermore, we partition $\R = \R_p \uplus \R_e \uplus \R_s$ into object roles that correspond to persistent objects, \eg machines or queue capacities, expected objects, \eg human resources that are present according to a specific schedule, and spontaneous objects, \eg patients or products that appear randomly.

We define three object roles for our delivery process example, namely packages $p$, deliverers $d$, and warehouse depots $w$, \ie $\R = \set{p, d, w}$, which are respectively spontaneous, expected and persistent objects. Furthermore, we define the following objects associated with these types: $\Ob_p = \mset{p_i \mid i \in \Nat^+}$, $\Ob_d = \mset{d_1, d_2}$, and $\Ob_w = \mset{2\cdot w_1, w_2}$. Note that while in reality, objects are normally distinguishable, which would imply that $\Ob$ is a set rather than a multiset. In practice, some objects do not have unique identifiers and we should consider $\Ob$ as a multiset, allowing for abstractions of objects. In the running example, spots in a depot are indistinguishable and we say that there are two $w_1$ objects to indicate that depot $w_1$ has space for two packages.

We define behavior $X = (\bar X, \prec_X)$ in the general sense as a partially ordered set (see Def.~\ref{def:poset}) of events as it allows for encoding sequences of events as well as parallelism while preserving determinism.

Behavior $X$ can be projected onto objects $O \leq \Ob$, which we write as $X\proj_O = (\set{(a,\min(O',O))\mid (a,O')\in \bar{X}, \supp(\min(O',O)) \neq \emptyset}, \prec_{X\proj_O})$, with $\prec_{X\proj_O}$ satisfying the condition that for each $(a_1,\min(O_1',O_1)),(a_2,\min(O_2',O_2)) \in \overline{X\proj_O}$, $(a_1,O'_1) \prec_X (a_2,O'_2) \iff (a_1,\min(O_1',O_1)) \prec_{X\proj_O} (a_2,\min(O_2',O_2))$.

For any object $o \in supp(\Ob)$, we also write $X\proj_o = X\proj_{\mset{o}}$, denoting a \emph{trace}.

This naturally extends to projections on roles $R \subseteq \R$, which we write as $X\proj_R = X\proj_{\Ob_R}$. Again for any $r \in \R$, we use $X\proj_r = X\proj_{\set{r}}$ as a shorthand notation.

\emph{System behavior} $S$ is the \emph{true} behavior of a process, where the involved objects are identifiable and the partial order is correct.

Through projections on individual objects, there is synchronization via the shared events, \ie for example with two objects $o_1, o_2 \in supp(\Ob)$, their behaviors $S\proj_{\set{o_1,o_2}}$ is synchronized via events in $\overline{S\proj_{\set{o_1,o_2}}} \setminus \overline{S\proj_{o_1}} \setminus \overline{S\proj_{o_2}}$.
Similarly, we see where objects from specific object roles synchronize through projection on object roles. 

\begin{figure}[tb]
\centering

\begin{subfigure}{0.99\textwidth}
  \centering
  \hspace*{-3.4cm}
  \includegraphics[scale=0.7]{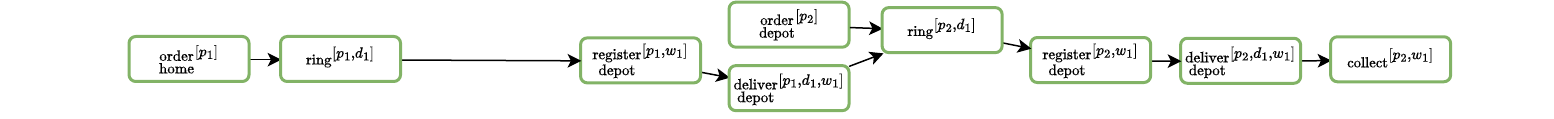}
  \caption{System behavior $S_1$.}
  \label{fig:S1}
\end{subfigure}
\begin{subfigure}{.99\textwidth}
  \centering
  \hspace*{-3.4cm}
  \includegraphics[scale=0.7]{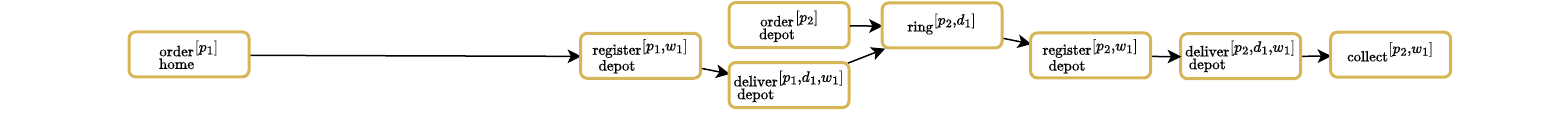}
  \caption{Recorded behavior, \ie system log, $L_1$.}
  \label{fig:L1}
\end{subfigure}
\begin{subfigure}{.99\textwidth}
  \centering
  \hspace*{-3.4cm}
  \includegraphics[scale=0.7]{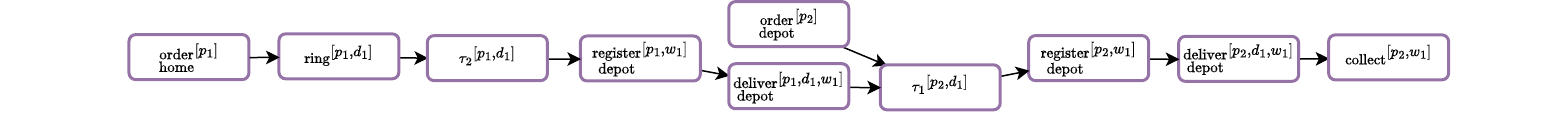}
  \caption{Process execution $\run_1 \in \Lang(M)$ of process model $M$.}
  \label{fig:run1}
\end{subfigure}
\caption{Visualization of the $S_1$, $L_1$, and $\run_1 \in \Lang(M)$ regarding packages 1 and 2.}
\label{fig:example1}
\end{figure}

Fig.~\ref{fig:S1} shows our first running example, where two packages are being processed for delivery, denoted as $S_1$. Package $p_1$ is ordered for home delivery, but is delivered at the nearest depot $w_1$ by deliverer $d_1$ after an unanswered door. In the meantime, package $p_2$ is ordered for depot delivery at $w_1$. However, deliverer $d_1$ tries to deliver it at home, denoted by the event $\ring{\mset{p_2,d_1}}$.

As the system is an `invisible' entity, the system behavior $S$ is considered an unknown poset of true events. Our goal is to reveal this behavior.

\subsection{System log: recorded behavior}\label{sec:log}
The execution of a process can be observed as a set of activity executions that are captured as recorded events. We write $e^\id = (a, t, O)^\id$ for a \emph{recorded event} $e^\id$ with activity $a$, timestamp $t$, and involved objects $O$. We omit the identifier and write $e$ instead of $e^\id$ when it is clear that $e^\id$ refers to an identifiable event. A logging mechanism in the system $S$ provides the recording of the events, either automatically or manually. In practice, there can be multiple logging mechanisms, usually characterized by object roles. For example, in the delivery process, the package orders and depot registrations are logged through a centralized system, whereas the deliverer performs manual logging for ringing and delivery activities. Lastly, the collection of packages is logged in the warehouse's system.

The order of the recorded events is based on their timestamps. As multiple logging mechanisms may have differently aligned clocks or record on different levels of granularity, the exact order between two events recorded in different mechanisms may be uncertain. Through synchronization of objects, we can treat the system log $L = (\bar L, \prec_L)$ as a single partially ordered set of recorded events, again omitting the timestamps, with partial order $\prec_L$ inferred in some way from the recording mechanism(s). This is also denoted as the \emph{recorded behavior}, therefore the object and role projections from Sec.~\ref{sec:system_behavior} are also defined on $L$.

The system log $L_1$ is the recorded behavior of $S_1$, depicted in Fig.~\ref{fig:L1}. Note that $L_1$ differs from $S_1$ as the first occurrence of ring, \ie $\ring{\mset{p_1,d_1}}$, is not recorded.

We assume the partial order is already inferred from knowledge about the recording mechanism(s). This is typically done through the timestamps when recording the events, and domain knowledge about the clocks' synchronization and abstraction level of the timestamp. Consider for example two clocks used to infer timestamps that are not synchronized, but we know that they will differ by more than 5 minutes.

\subsection{Modeled behavior}\label{sec:model}
Process models are frequently employed to describe and analyze the execution of a process. They can be manually created using domain knowledge or automatically discovered from the system’s recorded behavior, \ie the system log. With extended semantics of workflows, resources, and data attributes can be integrated into the process model, adding constraints on the activities they are involved in.

We emphasize that for the techniques we propose in this paper, any formalism of process models can be used to represent the behavior of $S$, provided its language can be defined by partial orders. This implies that $M$ is not required to be sound, but at least easy sound, \ie there is a process execution from the initial to the final state of $M$. Resource-constrained (RC) \nunets~\cite{sommers2022aligning} and Typed Jackson nets (t-JNs)~\cite{van2022data} are soundness-preserving subclasses of typed Petri nets with identifiers (t-PNIDS)~\cite{van2009generation,polyvyanyy2019information,van2022data}, able to incorporate the behavior of multiple interacting object roles. Alignment techniques that we developed initially specifically for RC \nunets~\cite{sommers2022aligning,sommers2023exact} naturally extend to t-JNs and easy sound t-PNIDs. While these examples of formalisms are restricted to modeling only one-to-one interactions between objects, other formalisms like Object-centric nets~\cite{van2020discovering} or synchronizing proclets~\cite{fahland2019describing} extend to a variable number of interacting objects, \ie with one-to-many and many-to-many relations.

\begin{figure}[tb]
\centering
\hspace*{-2.5cm}
\includegraphics[scale=0.6]{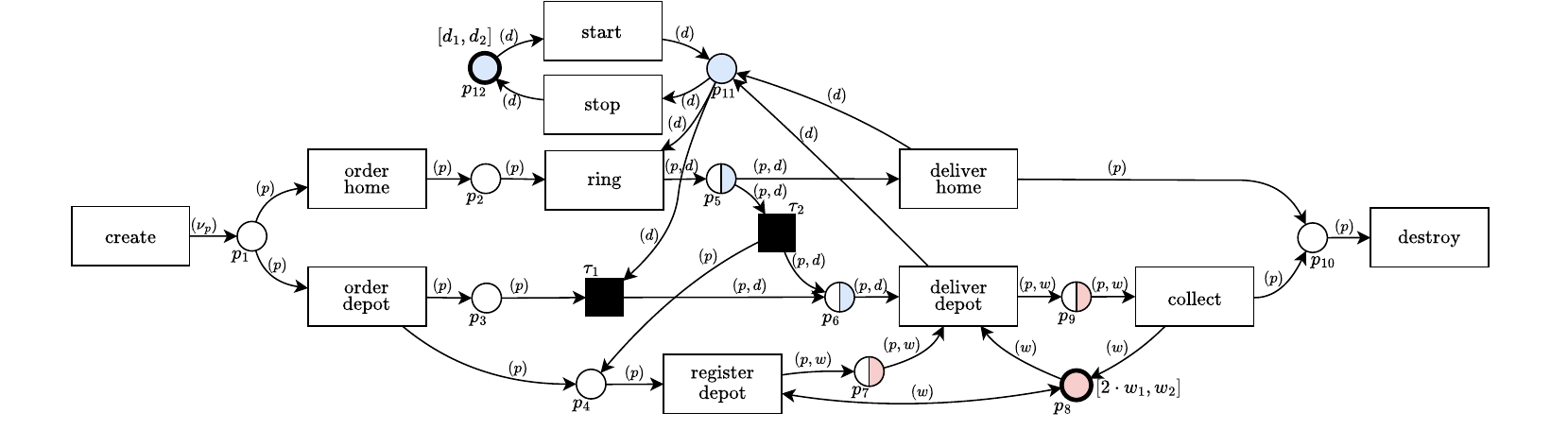}
    \caption{Running example $M = (N, m_i, m_f)$ with t-PNID $N$ and initial and final markings $m_i=m_f$. Furthermore, $M$ contains object roles $\R = \set{p,d,w}$ and variable sets $\V_p = \set{\nu_p,p},\V_d = \set{d}$, and $\V_w = \set{w}$.}
    \label{fig:re_net}
\end{figure}

To provide a visual depiction of the activity workflow for objects (packages, deliverers, and warehouse capacity) in the running example, we choose the semantics of t-PNIDs~\cite{van2022data}. Fig.~\ref{fig:re_net} shows the t-PNID $N = (P,T,F,\ell,\alpha,\beta)$ modeling the behavior of the package delivery process from our running example, with places $P$, transitions $T$, arcs $F \subseteq (P \times T) \cup (T \times P)$, and functions for transitions labels $\ell: T \rightarrow \Sigma$, place types $\alpha: P \rightarrow \R^*$, and arc variables $\beta: F \rightarrow (\V^*)^\oplus$. $N$ shows the choice from place $p_1 \in P$ between orders for home and depot delivery ($\orderhome{} \in T, \orderdepot{} \in T$). $\orderhome{}$ is followed by a $\ring{}$ activity which includes a deliverer object consumed from $p_{11}$. Either the door is answered and the package can be handed over ($\deliverhome{}$) or continues with silent transition $\tau_2$, $\registerdepot{}$, $\deliverdepot{}$, and $\collect{}$. After $\orderdepot{}$, there is a parallel split to claim a deliverer in $\tau_1$ and register the depot ($\registerdepot{}$) for delivery and collection at a specific depot. Packages are created and destroyed via $\create{}$ and $\destroy{}$ showing their limited lifetime. Deliverers are scheduled objects residing in place $p_{12}$ which can initiate and complete their availability through $\tstart{}$ and $\tstop{}$. Lastly, the persistent warehouse objects reside in place $p_8$.
The places' colors and the arcs' labels denote their types and variables, as defined by $\alpha$ and $\beta$ respectively. For its formal definition, we refer to~\cite{van2022data}. Variables are used to synchronize objects, while places with multiple types allow for encoding correlations between objects. \Eg variable $p$ on arcs $(p_6, \deliverdepot{})$ and $(p_7, \deliverdepot{})$ ensures that $\deliverdepot{}$ can only be executed when the package objects residing in $p_6$ and $p_7$ are the same. $p_7$ stores the correlation between packages and warehouses, ensuring that their interaction for $\deliverdepot{}$ is the same as in $\registerdepot{}$. We write $\Var(t)$ for the set of variables attached to transition $t$ and its size specifies the number of involved objects in $t$.

$N$ is completed with an initial and final state $m_i$ and $m_f$ to form the process model $M = (N, m_i, m_f)$. A t-PNID process model $M$ simultaneously models the behavior of multiple objects and their interaction with each other. Within $M$, each object behaves according to its role, which is defined by the place typing function $\alpha$, and is constrained by the behaviors of other objects through synchronization and correlations. Objects' modeled behavior can be isolated by projections of the process model. For some set of roles $R \subseteq \R$, we denote $N\proj_{\Ob_R}$ to be the projection of a net $N$ onto objects $\Ob_R$, and it is defined by the places in the net whose types intersect with $R$ and the corresponding arcs and transitions. For shorthand notation, we also write $N\proj_R$ for $N\proj_{\Ob_R}$, and $N\proj_o$ and $N\proj_r$ for $N\proj_{\mset{o}}$ and $N\proj_{\set{r}}$ for any $o \in supp(\Ob)$ and $r \in \R$ respectively. Similarly, projections are also defined on markings, denoting the state of a subset of objects, and therefore also on process models, combining nets and markings. For a formal definition of t-PNID projections, we refer to~\cite{sommers2024conformance}.

\begin{figure}[tb]
\centering
\begin{subfigure}{1\textwidth}
  \centering
  \hspace*{-1.9cm}
  \includegraphics[scale=0.6]{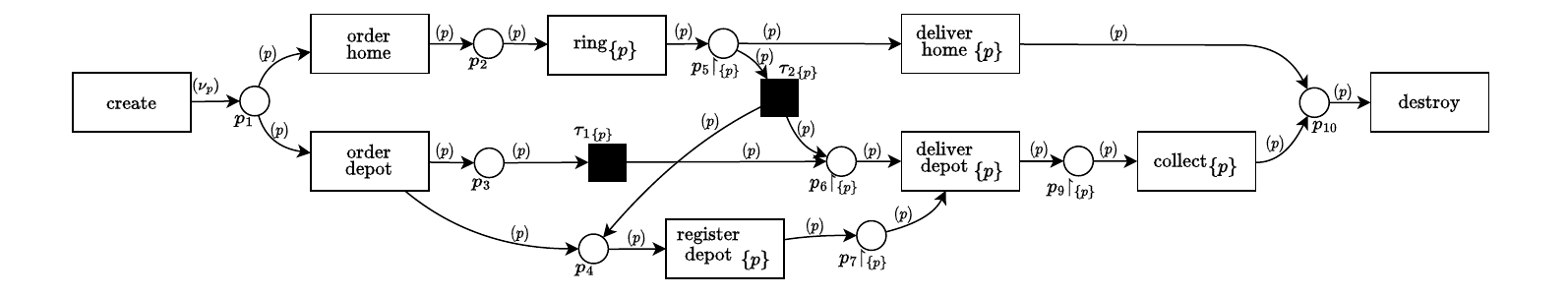}
  \caption{$M\proj_{\set{p}}$}
  \label{fig:re_net_p}
\end{subfigure}
\begin{subfigure}{.49\textwidth}
  \centering
  \hspace*{-1.4cm}
  \includegraphics[scale=0.6]{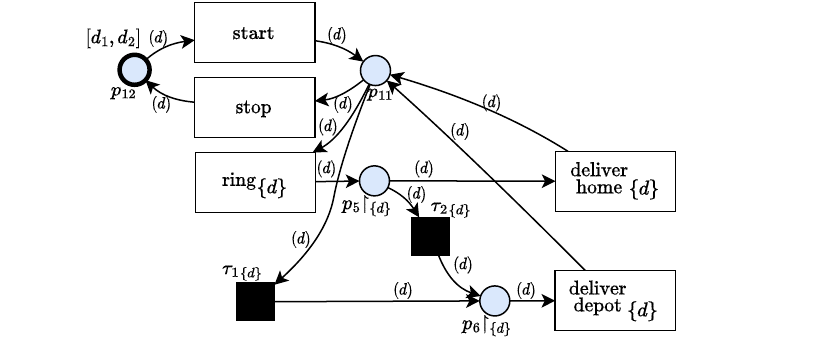}
  \caption{$M\proj_{\set{d}}$}
  \label{fig:re_net_d}
\end{subfigure}
\begin{subfigure}{.49\textwidth}
  \centering
  \hspace*{-1.6cm}
  \includegraphics[scale=0.6]{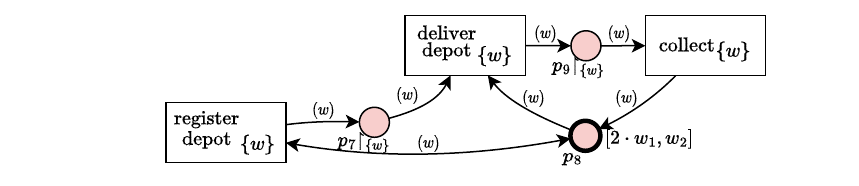}
  \caption{$M\proj_{\set{w}}$}
  \label{fig:re_net_w}
\end{subfigure}
\caption{Projections of $M$ on object types $p$, $d$, and $w$.} \label{fig:re_net_projections}
\end{figure}

The projections of $M$ on individual roles of packages $p$, deliverers $d$, and warehouses $w$ are shown respectively in Figs.~\ref{fig:re_net_p} to~\ref{fig:re_net_w}. We highlight that any interaction with other objects is ignored in the projection nets, and places and transitions that change properties are denoted with additional projection subscripts. \Eg transition $\ring{}\proj_{\set{p}}$ in $M\proj_{\set{p}}$ denotes a change in involved objects, and similarly place $p_5\proj_{\set{p}}$ denotes a change in correlated objects in its residing tokens.

The behavior of a process model $M$ is defined by its language $\Lang(M)$, denoting in the general sense its set of process executions. Note that this allows for any process model formalism for which the language is defined as execution posets of atomic activity executions. 

For t-PNIDs, such an atomic activity execution is defined by a \emph{transition firing} $t_\mode^\id$ on transition $t \in T$ for activity $\ell(t) \in \Sigma$ and involved objects $O \leq \Ob$ defined by the mode $\mode$. $\mode$ maps variables from the transition's incoming and outgoing arcs to object names defining the tokens that are consumed and produced by the transition firing. When it is clear that $t_\mode^\id$ refers to an identifiable transition firing, we again omit the identifier and write $t_\mode$ instead, for clarity. For example in $M$, initially only $\tstart{}$ and $\create{}$ are enabled respectively with modes $\mode(d) = d_1$ or $\mode(d) = d_2$, and $\mode(\nu_p) \in \V_p$. Firing the former results in a state change to marking $m$ with $m(p_{12}) = \mset{d_2}$ and $m(p_{11}) = \mset{d_1}$.

An \emph{execution poset} $\run = (\bar\run, \prec_\run)$ is a partially ordered set of transition firings where each totally ordered permutation is a firing sequence, \ie $m_i \xrightarrow{\run} m_f \iff \forall_{\sigma \in \run_M^>} m_i \xrightarrow{\sigma} m_f$.

Figure~\ref{fig:run1} depicts a process execution $\run_1 \in \Lang(M)$ that most closely matches the behavior of $S_1$. Note that any process execution from the language of $M$ can be used for comparing modeled behavior to recorded or system behavior. Next, we go into detail of such comparison, and later in Sect.~\ref{sec:alignments}, we show how the process execution is chosen that provides the most information into the system behavior.

\section{Behavioral deviations}\label{sec:deviations}
A system log $L$ generated by system $S$ shows the activities executed by $S$. However, most real-life system logs tend to be \emph{incomplete}, \emph{noisy}, and \emph{imprecise}, as logging mechanisms may be faulty and the integration of event data from different systems may be imprecise~\cite{bose2013wanna}. Manual logging imposes additional data quality risks.

A process model $M$ is modeled to represent the behavior $S$ of the system, either manually or automatically. However, manual modeling imposes quality risks when domain knowledge is incomplete and automatically discovering the model inherits the quality issues from $L$, making $M$ a possibly imprecise representation of $S$ as well.

To compare poset representations, and reveal these imprecisions in $L$ and $M$, we first define the possible matching of the events in the different representations to separate conformities and deviations. We concretize these deviations with regard to the real behavior of $S$.

\subsection{Comparing behavioral representations}\label{sec:compare}
The system $S$, log $L$, and executions of process model $M$ all reason over the behavior of the system, in true, recorded, and modeled sense, each having a different domain. Where $S$ reasons about events, $L$ and $M$ reason about recorded events and transition firings respectively, both referring to events in $S$. In order to expose conformities and deviations, we define a behavioral congruence relation that respects a matching function, which decides whether two elements could be referring to the same event, and the partial orders of the posets. In the case of an element not being congruent with any element in the other representation, we speak of deviations, for which we introduce an additional element $\varepsilon$.
\begin{definition}{(Matching, Behavioral congruence)}\label{def:match}
    Let $X = (\bar X, \prec_X)$ and $Y = (\bar Y, \prec_Y)$ be two partially ordered sets. We extend its elements with $\varepsilon$, denoted by $\bar{X}_\varepsilon = \bar{X} \uplus \set{\varepsilon_X}$ and $\bar{Y}_\varepsilon = \bar{Y} \uplus \set{\varepsilon_Y}$.

    $x \sim y$ denotes that elements $x \in \bar X$ and $y \in \bar Y$ are \emph{potential matches} according to a \emph{matching function} $\match: \bar X_\varepsilon \times \bar Y_\varepsilon \rightarrow [0,1]$, \ie $x \sim y \iff \match(x,y) \geq \theta$ for some $\match$ and $0 \leq \theta \leq 1$. For any $x \in \bar X$ and $y \in \bar Y$, we say that $\match(x,\varepsilon_Y) = 1$ and $\match(\varepsilon_X, y) = 1$, and therefore $x \sim \varepsilon_Y$ and $\varepsilon_X \sim y$.

    Furthermore, $\cong \subseteq \sim$ denotes an instance of $\sim$ that respects the original partial order $\prec_X$ and $\prec_Y$ and each element of $\bar X$ and $\bar Y$ are congruent with exactly one of $\bar X_\varepsilon$ and $\bar Y_\varepsilon$ respectively, \ie
    \begin{gather}
        \forall_{x \in \bar X} \exists!_{y \in \bar{Y}_\varepsilon} x \cong y \text{ and } \forall_{y \in \bar Y} \exists!_{x \in \bar{X}_\varepsilon} x \cong y \\
        \not\exists_{(x_1,y_1),(x_2,y_2) \in \cong \cap (\bar X \times \bar Y)} (x_1 \prec_X x_2) \wedge (y_2 \prec_Y y_1)
    \end{gather}
\end{definition}
\begin{figure}[tb]
\centering
\begin{subfigure}{.32\textwidth}
  \centering
  \includegraphics[scale=0.8]{Figure2.pdf}
  \caption{Poset $A$.}
  \label{fig:poset_A}
\end{subfigure}
\begin{subfigure}{.32\textwidth}
  \centering
  \includegraphics[scale=0.8]{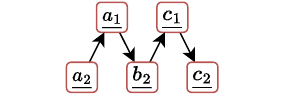}
  \caption{Poset $B$.}
  \label{fig:poset_B}
\end{subfigure}
\begin{subfigure}{.32\textwidth}
  \centering
  \includegraphics[scale=0.8]{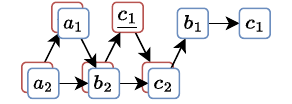}
  \caption{Congruence relation $\cong$ between $A$ and $B$.}
  \label{fig:congruence_AB}
\end{subfigure}
\caption{Examples of posets and the congruence relation.} \label{fig:poset_congruence}
\end{figure}
Two elements $x \in \bar X$ and $y \in \bar Y$ are potential matches, \ie $x \sim y$, if they could concern the same activity and the same involved objects. For example, with $X = L$ and $Y = \run \in \Lang(M)$, a recorded event $(a, O) \in \bar L$ and a transition firing $t_\mode \in \run$ are potential matches if $\ell(t) = a$ and $\obj(t_\mode) = O$. Note that with multiple transition firings with the same $t$ and $\mode$, there are more potential matches.

The congruence relation takes a one-to-one mapping from $\sim$ that also takes into account the partial order, such that the two partial orders may not disagree on the ordering of matched elements. Take for example the two posets illustrated in Figs.~\ref{fig:poset_A} and~\ref{fig:poset_B}. With $\sim = \set{ (a_1, \underline{a_1}), (c_1, \underline{c_1}), (a_2, \underline{a_2}), (b_2, \underline{b_2}), (c_2, \underline{c_2}) }$, we have the congruence relation as depicted in Fig.~\ref{fig:poset_congruence}. $(c_1, \underline{c_1}) \notin \cong$ since they disagree on the ordering of the matched elements, \ie $c_2 \prec_A c_1$ and $\underline{c_1} \prec_B \underline{c_2}$. Furthermore, there is no element in $B$ to which $b_1 \in A$ matches. Therefore, $(\varepsilon_A, \underline{c_1}), (c_1, \varepsilon_B)$, and $(b_1, \varepsilon_B)$ denote the differences or disagreements between the two posets according to the $\cong_{A,B}$, which we call \emph{behavioral deviations} when applied on behavioral representations:
\begin{definition}{(Behavioral deviations)}\label{def:deviation}
Let $X$ and $Y$ be two behavioral representations and $\cong$ a congruence relation between $X$ and $Y$.

A \emph{behavioral deviation} is an element $(x, y) \in \cong$ with either $x = \varepsilon_X$ or $y = \varepsilon_Y$. $(x, \varepsilon_Y)$ denotes either an incorrect event in $X$ with regard to $Y$ or a missing event in $Y$ with regard to $X$. Similarly, $(\varepsilon_X, y)$ denotes either an incorrect event in $Y$ with regard to $X$ or a missing event in $X$ with regard to $Y$.
\end{definition}
Our goal is to maximize the information from the recorded and the modeled behavior to classify these deviations regarding the system behavior. By doing so, we aim to understand which behavior most likely happened in reality and which stems from either recording or modeling imprecisions. The classification could be based on the trust of the representations in question. For example, with full trust in the recorded behavior, each deviation between $L$ and $M$ denotes a mistake in the model. Later we go into more detail on this and propose a method that is independent of the assumed trust levels. To do so, we first go into detail on how these deviations in recorded and modeled behavior relate to the system, \ie reality.

\subsection{Behavioral deviations with regard to reality}\label{sec:issues}
We concretize the behavioral deviations from Def.~\ref{def:deviation} through data quality issues as they have been categorized into different classes of problems across various entities, as shown in Table~\ref{tab:L_issues}. These issues have been categorized for recorded behavior $L$, which we generalize here to modeled behavior $M$ as well, by discussing how each of these quality issues can be found both in the log and the model.
\begin{table}[tb]
\centering
\captionsetup{width=\linewidth}
\caption{Categorization of data quality issues for each entity in recorded behavior, taken from~\cite{bose2013wanna}. $S$, $L$, and $\run \in \Lang(M)$ denote respectively the true behavior, the recorded behavior, and the modeled behavior represented by \emph{a} process execution in the process model $M$.}\label{tab:L_issues}
\begin{tabular}{|l|r|r|r|r|r|r|r|r|r|}
\hline
    & \multicolumn{1}{c|}{\rotatebox{90}{\bf{case}}} & \multicolumn{1}{c|}{\rotatebox{90}{\bf{event}}} & \multicolumn{1}{c|}{\rotatebox{90}{\bf{belongs to}}} & \multicolumn{1}{c|}{\rotatebox{90}{\bf{c\_attribute}}} & \multicolumn{1}{c|}{\rotatebox{90}{\bf{position}}} & \multicolumn{1}{c|}{\rotatebox{90}{\bf{activity name~}}} & \multicolumn{1}{c|}{\rotatebox{90}{\bf{timestamp}}} & \multicolumn{1}{c|}{\rotatebox{90}{\bf{resource}}} & \multicolumn{1}{c|}{\rotatebox{90}{\bf{e\_attribute}}} \\ \hline
Missing data ($S \setminus L$ or $S \setminus \run$)   & & \Lissue{mi}{e}  & \Lissue{mi}{o}   &  & \Lissue{mi}{p}   &  &  & \Lissue{mi}{o}   & 
\\ \hline
Incorrect data ($L \setminus S$ or $\run \setminus S$) &  & \Lissue{in}{e}  & \Lissue{in}{o}   &   & \Lissue{in}{p}   &   &    & \Lissue{in}{o}   & 
\\ \hline
Imprecise data  & & &    &    &   &    &  &  &   \\ \hline
Irrelevant data & &   &&&&&&&   \\ \hline
\end{tabular}
\end{table}

In this paper, we focus on the categories of \emph{missing} and \emph{incorrect} data, and disregard data that is \emph{irrelevant} for the process. We deem granularity issues in \emph{imprecise} data as a separate problem, and we do not address it here. 
In terms of the data entities, we only use the events' timestamps to derive the partial order relation, and assume that the \emph{activity name} and \emph{belongs to} attribute (object id)  are never missing. This leaves us with the following data quality issues, which we explain with the system $S$, with system log $L$, and \emph{a} process execution $\run \in \Lang(M)$, and additionally concrete examples from the package delivery process.
\begin{itemize}
    \item[\Lissue{mi}{e}] An event that has occurred in reality is missing within $L$ or $\run$, \ie either the recorded or modeled behavior. Such missing behavior is contained in $S \setminus L$ and $S \setminus \run$. For example, with $S_1$, $L_1$, and $\run_1$ from Fig.~\ref{fig:congruence1}, we see a missing recorded event in $\stup{\varepsilon_{L_1}, \ring{\mset{p_1,d_1}}} \in \cong_{L_1,S_1}$ as it was not logged, and a missing transition firing in $\stup{\varepsilon_{\run_1}, \ring{\mset{p_2,d_1}}} \in \cong_{\run_1,S_1}$ because $M$ does not allow for $\orderdepot{}$ and $\ring{}$ to be present in the same process execution.
    \begin{figure}[tb]
        \centering
        \hspace*{-3.cm}
        \includegraphics[scale=0.7]{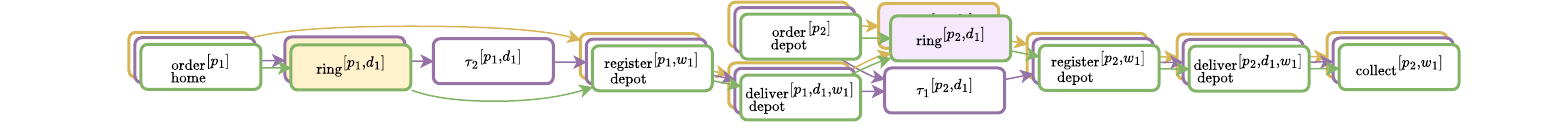}
        \hspace*{3.cm}
        \caption{Congruence relation $\cong_{L_1,\run_1,S_1}$ between $L_1, \run_1 \in \Lang(M)$, and $S_1$.}
        \label{fig:congruence1}
    \end{figure}
    \item[\Lissue{in}{e}] An event is logged or modeled incorrectly. This means that an event that is not executed in reality is recorded in the log or is included in the process execution of $M$, \ie the event is in $L \setminus S$ or $\run \setminus S$. Consider for example $S_1$, $L_1$, and $\run_1$ from Fig.~\ref{fig:congruence2}. We see that for package $p_3$, $\ring{}$ was logged incorrectly as it did not occur, \ie $\stup{\ring{\mset{p_3,d_1}},\varepsilon_{S_2}} \in \cong_{L_2,S_2}$. For package $p_4$, we see that $\ring{}$ was skipped, while the model requires a move on the $\ring{}$ transition, \ie $\stup{\ring{\mset{p_4,d_1}}, \varepsilon_{S_2}} \in \cong_{\run_2, S_2}$. Package $p_5$ is added to show that the deliverer was occupied during the timeframes where $\ring{\mset{p_3,d_1}}$ and $\ring{\mset{p_4,d_1}}$ are supposed to have happened according to $L_1$ and $M$ respectively.
    \begin{figure}[tb]
        \centering
        \hspace*{-2.2cm}
        \includegraphics[scale=0.7]{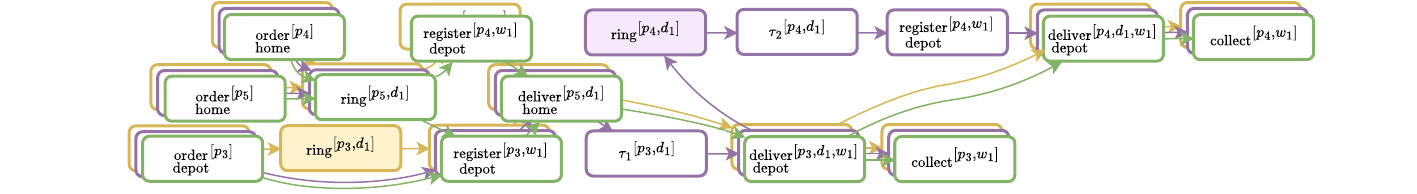}
        \hspace*{2.2cm}
        \caption{System log $L_2$, run $\run_2 \in \Lang(M)$, and system $S_2$, and their congruence relation $\cong_{L_2,\run_2,S_2}$. 
        }
        \label{fig:congruence2}
    \end{figure}
    \item[\Lissue{mi}{o}]
    An event misses information about an involved object, which belongs to $S \setminus L$ or $S \setminus \run$. For example, in the scenario where a deliverer ($d_2$) delivers a package to a different depot than is registered for. They may not want to be associated with this deliberate deviation from the instructions and log the activity without their name, as depicted in $\stup{\varepsilon_{L_3}, \deliverdepot{\mset{p_6,d_2,w_1}}}, \stup{\deliverdepot{\mset{p_6,w_2}}, \varepsilon_{S_3}} \in \cong_{L_3,S_3}$ in Fig.~\ref{fig:congruence3}. For this quality issue to occur in the modeled behavior, an object role would be missing in $M$, \eg the workflow from the perspective of the delivery truck or even the client, not included in our running example.
    \item[\Lissue{in}{o}]
    The association between an event and an involved object is logged or modeled incorrectly, again corresponding to behavior in $L \setminus S$ or $\run \setminus S$. Consider again package $p_6$, which was automatically registered to warehouse depot $w_2$, however, it is delivered to $w_1$ by deliverer $d_2$, even though $d_2$ logs the delivery to $w_2$ to conceal the mistake, as can be seen by $\stup{\varepsilon_{L_3}, \deliverdepot{\mset{p_6,d_2,w_1}}}, \stup{\deliverdepot{\mset{p_6,w_2}}, \varepsilon_{S_3}}\allowbreak \in \cong_{L_3,S_3}$. Since the package is collected at $w_1$, it is evidently a recording error. On the other hand, package $p_7$ is being handed over from deliverer $d_1$ to $d_2$ for depot delivery as $d_1$'s shift is ending. The process model $M$ does not allow for a handover, resulting in a mismatch seen in $\stup{\varepsilon_{\run_3}, \deliverdepot{\mset{p_7,d_2,w_1}}} \in \cong_{\run_3,S_3}$, $\stup{\deliverdepot{\mset{p_7,d_1,w_1}}, \varepsilon_{S_3}}\allowbreak \in \cong_{\run_3,S_3}$, and $\stup{\varepsilon_{\run_3}, \tstop{\mset{d_1}}}, \stup{\tstop{\mset{d_1}}, \varepsilon_{S_3}} \in \cong_{\run_3,S_3}$.
    \begin{figure}[tb]
        \centering
        \hspace*{-1.2cm}
        \includegraphics[scale=0.7]{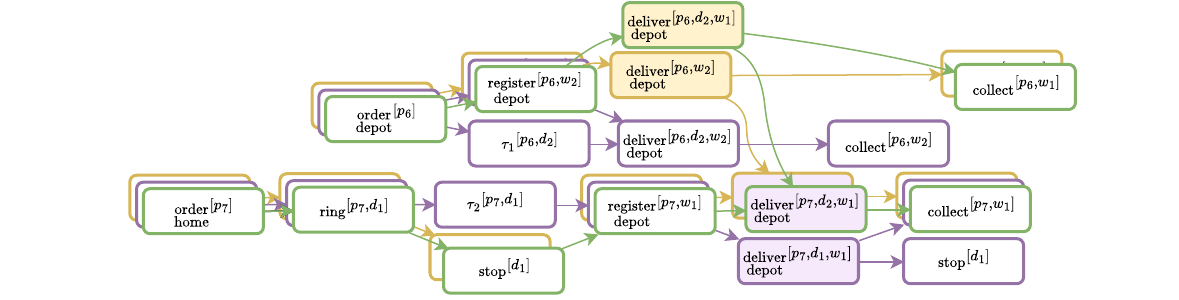}
        \hspace*{1.2cm}
        \caption{System log $L_3$, run $\run_3 \in \Lang(M)$, and system $S_3$, and their congruence relation $\cong_{L_3,\run_3,S_3}$. 
        }
        \label{fig:congruence3}
    \end{figure}
    \item[\Lissue{mi}{p}]
    With system logs as partially ordered sets, there may be events for which the exact time information is missing, so their exact order cannot be established. The events appear in parallel, although their actual execution was ordered ($S \setminus L$). Similarly, process executions in $M$ only capture ordering information between events concerning the same objects, which can cause information loss ($S \setminus \run$).
    \item[\Lissue{in}{p}]
    Events in the log may be ordered differently than in reality, \eg in a  situation similar to the one in \Lissue{mi}{p}, the events may be recorded too late after some other events occurred and were timely recorded in the log. Similarly, the model may impose ordering restrictions on objects that are violated in $S$. This behavior is reflected in $L \setminus S$ and $S \setminus L$ respectively, since the log or the model includes information not matching the reality, and misses some information that should have been captured. Fig.~\ref{fig:congruence4} shows the scenario where $\ring{\mset{p_8,d_1}}$ is recorded after it occurred, \ie $\stup{\ring{\mset{p_8,d_1}}, \varepsilon_{S_4}}, \stup{\varepsilon_{L_4}, \ring{\mset{p_8,d_1}}} \in \cong_{L_4,S_4}$, due to the nature of manual logging, and the scenario where $d_1$ was multitasking. \Ie they ring for package $p_9$ and before the delivery move to the next house to ring and deliver package $p_{10}$. This is ordering is not allowed in $M$, hence we see $\stup{\varepsilon_{\run_4}, \ring{\mset{p_{10},d_1}}}$ and $\stup{\ring{\mset{p_{10},d_1}}, \varepsilon_{S_4}}$ in $\cong_{\run_4,S_4}$.
    \begin{figure}[tb]
        \centering
        \hspace*{-1.4cm}
        \includegraphics[scale=0.7]{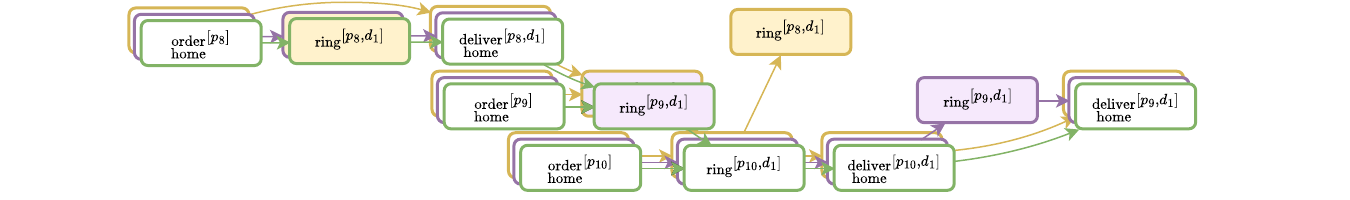}
        \hspace*{1.4cm}
        \caption{System log $L_4$, run $\run_4 \in \Lang(M)$, and system $S_4$, and their congruence relation $\cong_{L_4,\run_4,S_4}$. 
        }
        \label{fig:congruence4}
    \end{figure}
\end{itemize}
Bose et al. \cite{bose2013wanna} provide a comprehensive detailed account of each data quality issue.

\section{Describing reality by alignments}\label{sec:alignments_repr}
\emph{Conformance checking} is a technique aimed at the comparison of the behavior of the model $M$ of system $S$ and the observed behavior of system $S$ as recorded in log $L$. Several state-of-the-art techniques in conformance checking use alignments to relate a recorded execution of a process to the most similar trace of the language of a process model~\cite{carmona2018conformance,van2012replaying}. More advanced techniques additionally incorporate data or resource information~\cite{alizadeh2018linking,de2013aligning,mannhardt2016balanced,mehr2021detecting}. An alignment exposes deviations of the log execution from this trace explaining where the activities prescribed by the process model were not executed and where activities not allowed by the model were performed. The techniques referred to above operate on a trace-by-trace basis, not incorporating inter-case dependencies from interacting objects, and thus failing to expose deviations on these. In previous work, we proposed a method to align a complete system log to a resource-constrained \nunet, which as described above is a subclass of t-PNIDs.

Alignments integrate information from both representations, the log, and the model, and give us an enriched representation of reality. This enriched representation is influenced by the assumption about the similarity measure between $L$ and $M$ and our confidence in $L$ and $M$ in relation to $S$, which we call trust. We show that these assumptions can cause ambiguity in the relations from $L$ and $M$ to $S$ and therefore obscure the trust. In the next section, we propose a method to remove some of the assumptions and therefore reduce this ambiguity and strengthen the trust in $L$ and $M$.

\subsection{Aligning recorded and modeled behavior}\label{sec:alignments}
Recall that the congruence relation from Def.~\ref{def:match} aligns two behaviors defining a matching relation of their elements while respectively the partial orders. At the core of alignments of recorded behavior $L$ and a process model $M$ are \emph{moves}, defining the matching function on recorded events and transition firings. Alignment moves belong to one of the following three types, which we formally define in Def.~\ref{def:align_moves}:
\begin{enumerate}
    \item \emph{log moves}, indicating that a recorded event cannot be mimicked by the process model;
    \item \emph{model moves}, indicating that the model requires the execution of an activity that has no matching recorded event at the corresponding position;
    \item \emph{synchronous moves}, indicating that observed and modeled behavior agree on the event represented by the move.
\end{enumerate}
Log and model moves refer to unmatched elements from respectively recorded and modeled behavior, which we denote as \emph{deviating moves}, while synchronous moves are \emph{conforming moves} referring to elements that are matched in both behaviors.

\begin{definition}{(Log, model and synchronous moves)}\label{def:align_moves}
Let $L$ be a system log and $M = (N, m_i, m_f)$ be a process model with t-PNID $N = (P, T, F, \ell, \alpha, \beta)$ and $T_\mode$ the set of all possible transition firings in $M$ (\cf Def.~\ref{def:marking}).

$\al_s = \set{(e, t_\mode) \mid e = (a, O) \in L, t_\mode \in T_\mode, a = \ell(t), O = \obj(t_\mode)}$ is the set of possible \emph{synchronous moves}, following the \emph{potential matching function} from Def.~\ref{def:match}. $\al_l = \set{(e, \gg) \mid e \in L}$ is the set of possible \emph{log moves}, and $\al_m = \set{(\gg, t_\mode)^\id \mid t_\mode \in T_\mode}$ is the infinite set of possible \emph{model moves}, with the additional identifier to make each move unique.
\end{definition}

An \emph{alignment} defines the congruence relation between $L$ and $M$, \ie a poset over the set of synchronous, log and model moves that incorporates both the recorded behavior and the allowed behavior of the model. The alignment's deviating moves expose deviation of the recorded behavior from the modeled behavior while the synchronous moves show where the recorded behavior follows the model.

\begin{definition}{(Alignment)}\label{def:alignment}
Let $L=(\bar{L},\prec_L)$ be a system log and $M = (N, m_i, m_f)$ be a process model. An \emph{alignment} $\al$ is a poset $\al=(\bar\al, \prec_\al)$, where $\bar\al \subseteq (\al_l \cup \al_m \cup \al_s)$, with the following properties:
\begin{enumerate}
    \item $\overline{\al\proj_L} = \bar{L}$ and $\prec_L \subseteq \prec_{\al\proj_L}$ \label{prop:align1}
    \item $m_i \xrightarrow{\al\proj_T} m_f$, \ie $\forall_{\sigma \in (\al\proj_T)^<}, m_i \xrightarrow{\sigma} m_f$ \label{prop:align2}
\end{enumerate}
with alignment projections on the log events $\al\proj_L$ and on the transition firings $\al\proj_T$:
\begin{align}
    \al\proj_L &= \left( \sset{e \smid \stup{e, t_\mode} \in \bar\al, e \neq \gg}, \sset{\stup{e,e'} \smid \stup{\stup{e, t_\mode}, \stup{e', t'_\mode}} \in \prec_\al, e \neq \gg \neq e'}  \right) \label{eq:logrun} \\
    \al\proj_T &= \left( \sset{t_\mode \smid \stup{e, t_\mode} \in \bar\al, t_\mode \neq \gg}, \sset{\stup{t_\mode, t'_\mode} \smid \stup{\stup{e, t_\mode}, \stup{e', t'_\mode}} \in \prec_\al, t_\mode \neq \gg t'_\mode} \right) \label{eq:modelrun}
\end{align}
\end{definition}
Log and model moves relate to deviations as defined in Def.~\ref{def:deviation}, and therefore the ambiguity in explanation as presented in Def.~\ref{def:deviation} transfers to these moves similarly. In Sec.~\ref{sec:issues}, we described how these deviations relate to the system, showing that both the model and the log may be imprecise. Hence, a model (or log) move could signify that something has happened but not logged (modeled), or it signifies that it is modeled (logged) incorrectly. We emphasize here that synchronous moves are defined atomically, meaning that they either synchronize the behavior from an event completely or not at all. From projections on objects, we can obtain alignments regarding only those objects. When the behavior from the perspective of a single involved object can not be synchronized, there exists no synchronous move for the recorded event and corresponding transition firings. Regarding the likelihood of recorded events and transition firings belonging to the system's behavior, we assume that it decreases when fewer involved objects allow synchronization.

Since the moves relate to elements in $\cong_{L,\run}$, Figs.~\ref{fig:congruence1} to~\ref{fig:congruence4}, depict the congruence relations as well as the alignments of $M$ and $L_1$ to $L_4$.

The core alignment question is then to find an execution poset $\run$ in $M$, \ie $m_i \xrightarrow{\run} m_f$ that matches most closely with $L$ (\cf Def.~\ref{def:align_moves}), \ie maximizing the number of synchronous moves, and minimizing the number of log and model moves. Various techniques exist for computing this optimal alignment, as discussed in~\cite{sommers2022aligning} and~\cite{sommers2023exact}.

\subsection{\texorpdfstring{$\al_{L,M}$}{An alignment} as a representation of \texorpdfstring{$S$}{reality}: a matter of trust}\label{sec:trust}
An alignment $\al$ is a representation of the system $S$ (\cf Fig.~\ref{fig:SLM}) as it integrates two behavioral representations $L$ and $M$ by selecting an execution poset $\run \in \Lang(M)$ and a congruence relation ($\cong_{L,\run}$) between the two, distinguishing where they agree and disagree. Conformities $(e, t_\mode) \in \cong$, and deviations $(e, \varepsilon_M) \in \cong$ and $(\varepsilon_L, t_\mode) \in \cong$ are denoted by synchronous moves ($\al\proj_{\al_s}$), log moves ($\al\proj_{\al_l}$), and model moves ($\al\proj_{\al_m}$) respectively. Even without deviations, \ie $\al = \al\proj_{\al_s}$, the alignment provides an enriched representation of $S$ by combining activity executions with structural properties of the process. Although individual trace alignments fail to capture inter-object dependencies present in the system, the complete system log alignment, introduced in~\cite{sommers2022aligning}, can already provide a more accurate representation of reality.

Assuming that the log and the model are the only sources of our knowledge about system $S$, we can say that to the best of our knowledge, synchronous moves are part of the behavior of $S$, as this is where $L$ and $\run$ agree, \ie for each $(e,t_\mode) \in \al$ there is exactly one event $s \in S$ such that $(e, t_\mode, s) \in \cong_{L,\run,S}$. However, for model and log moves it is not trivial to decide whether they represent system behavior. An alignment is considered to be optimal if it minimizes the cost associated with the moves, without taking into account potential data quality issues, \ie with a cost function $c: \al \rightarrow \mathbb{R}^+$, alignment $\al$ is optimal such that $\sum_{g \in \al} c(g) \leq \sum_{g \in \al'} c(g)$ holds for any alignment $\al'$. Generally, the cost function is defined such that the number of synchronous moves is maximized and the number of log and model moves are minimized, which can be achieved by defining the \emph{standard} cost function for a move $(e, t_\mode) \in \al$  as follows.
\begin{equation}\label{eq:cost}
    c\stup{\stup{e, t_\mode}} =
    \begin{cases}
        1        & (e, t_\mode) \in \al_{lm} \wedge \ell(t) \neq \tau \\
        0        & \text{otherwise\footnotemark}
    \end{cases}
\end{equation}
\footnotetext{In theory we should have $c(t) > 0$ when $\ell(t) = \tau$, since otherwise it may cause problems when there is an infinite loop behavior possible by firing only invisible transitions~\cite{adriansyah2014aligning}.}

Trust levels $\tau_M, \tau_L$ are metrics for model and log quality, respectively, and they refer to the degree to which a model/log is reliable and accurate~\cite{rogge2016log}. Trust levels take values from 0 to 1. 
In Tab.~\ref{tab:trusts}, we show how different trust levels for $L$ and $M$ influence the representation of reality based on an alignment. A deviation $(e, \gg) \in \al\proj_{\al_l}$ or $(\gg, t_\mode) \in \al\proj_{\al_m}$ exposed in $\al$ can be interpreted as issues to fix in either of the representations, denoted as log and model repair~\cite{fahland2015model, rogge2014temporal, rogge2013improving, wang2015cleaning}. Note that with $\tau_M \neq \tau_L$, it is not necessarily desirable to optimize for the minimization of model and log moves, but prefer one over the other. This can easily be achieved by modifying the cost function $c$ such that $c((e, \gg)) \neq c((\gg, t_\mode))$ for some log move $(e, \gg) \in \al_l$ and model move $(\gg, t_\mode) \in \al_m$.
\begin{table}[tb]
\caption{Different trusts and respective (in)correct classifications.}\label{tab:trusts}
\begin{tabular*}{\textwidth}{@{\extracolsep\fill}lcccccc}
\end{tabular*}
\centering
\hspace*{-0.5cm}
\includegraphics[scale=0.7]{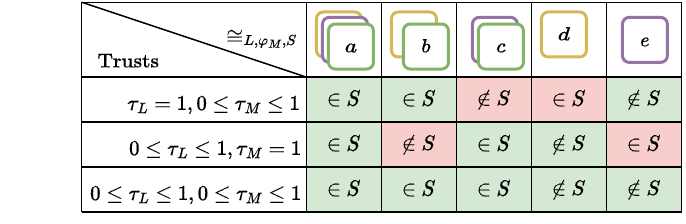}
\hspace*{0.5cm}
\end{table}

With complete trust in the process model, \ie $\tau_M = 1$ while $0 \leq \tau_L \leq 1$, the alignment $\al$ defines a congruence relation $\cong_{L,\run}$ which represents $S$ by the synchronous and model moves but not by the log moves. With this assumption, $\cong_{L,\run,S}$ only contains elements of the form $(e,t_\mode, s), (\varepsilon_L,t_\mode, s)$ and $(e, \varepsilon_\run, \varepsilon_S)$, as depicted in row 1 of Tab.~\ref{tab:trusts} for events $a, c$ and $d$.
When log moves are not considered part of the $S$, we generally modify the standard cost function such that $0 < c((\gg, t_\mode)) < c((e, \gg))$ for any ${(\gg, t_\mode)} \in \al_m$ and $(e, \gg) \in \al_l$. This is typically done in the settings of decision point analysis, where statistics on decision points in the process model are gathered~\cite{de2013data}, and log repair, where data quality issues are classified for repair~\cite{rogge2014temporal,rogge2013improving,wang2015cleaning}.

On the other hand, with complete trust in the system log, \ie $\tau_L = 1$, and $0 \leq \tau_M \leq 1$, only the synchronous and log moves of the alignment $\al$ represent $S$. With this assumption, $\cong_{L,\run,S}$ only contains elements of the form $(e,t_\mode, s), (e,\varepsilon_L,s)$ and $(\varepsilon_L,t_\mode,\varepsilon_S)$, as depicted in row 2 of Tab.~\ref{tab:trusts} for the events $a, b$ and $e$.
Similarly, we modify the cost function to prefer log moves over model moves, \ie $0 < {c((e, \gg))} < c((\gg, t_\mode))$ for any $(e, \gg) \in \al_l$ and $(\gg, t_\mode) \in \al_m$. The system log is typically trusted over the process model in settings of performance analysis, where statistics of the activity executions in the process are gathered~\cite{van2011process}, and of model repair, where the focus is on finding a better fitting process model based on a system log~\cite{fahland2015model}. 

If we cannot have complete trust in either of the two representations, \ie $0 \leq \tau_L, \tau_M \leq 1$, interpretations shown in row 1 and 2 of Tab.~\ref{tab:trusts} and modified cost functions cannot be used, since they would lead to interpretation mistakes. The green and red colored cells in Tab.~\ref{tab:trusts} indicate which behavior is misclassified due to invalid trust.
Ideally, one would need to correctly classify each deviation separately in order to get to the optimal representation of $S$ from $\run$ and $L$, shown in row 3 of Tab.~\ref{tab:trusts}, \ie $\cong_{L,\run,S}$ could contain a combination of the elements described before, represented by events $a$ to $e$ in Tab.~\ref{tab:trusts}.

Next, we go into detail on how the alignment $\al$ between $L$ and $M$ is computed which describes the congruence relation $\cong_{L,\run}$ between the two representations.

\section{Classifying deviations by relaxations}\label{sec:relaxations}
In this section, we propose a method to relax the alignment problem that allows for partial matching of recorded events and transition firings. We discuss how this improves the interpretability of deviating behavior by showing more accurately for which objects the recorded and modeled behavior agree. Exposing the deviations for different perspectives can provide information for the trust levels for both the process model and the system log locally, as discussed in Sec.~\ref{sec:trust}.

In Sec.~\ref{sec:issues}, we showed, through examples, the limitations of regular alignments with regard to the interpretability of deviations from known quality issues in both the recorded and modeled behavior. Here we revisit these examples and show how our proposed relaxed alignment extracts more information from the model and the log from which we can reveal which behavior would best match that of the system.

\subsection{Partial matching of log and model}\label{sec:partial}
A process combines multiple perspectives from the various object roles, \eg case, and resource perspectives. The modeled behavior, defined by $M = (N, m_i, m_f)$ with $N = (P, T, F, \ell, \alpha, \beta)$ distinguishes between these object roles through its place type function $\alpha$. $M$ defines the combination of the process models for each object role by synchronization of objects in transition firings. On the other hand, $L$ combines traces, denoting projections of the system log on the individual objects, that are synchronized in events involving multiple objects.

Recall from the previous section that $M$ and $L$ are compared in their entirety through an alignment $\al$, by defining a potential matching relation $\sim$ and computing a congruence relation $\cong$ from the alignment $\al$. The alignment $\al$ is defined (\cf Def.~\ref{def:alignment}) in such a way that any two elements $(a, O) \in \bar{L}$ and $t_\mode \in \run \in \Lang(M)$, with $\run = \al\proj_T$, can only match, and therefore be congruent, if they agree on all involved objects simultaneously. When for a single object $o \in (\supp(O) \cup \obj(t_\mode))$ the recorded and modeled behavior does not match, we have $(a, O) \not\cong t_\mode$, resulting in a log move $((a,O),\varepsilon_M) \in \cong$ or model move $(\varepsilon_L,t_\mode) \in \cong$. We discussed in Sec.~\ref{sec:issues} how these deviations are ambiguous in the sense of whether they have actually occurred, \ie whether they are included in the behavior $S$ or not.

We argue that, in general, the more involved objects allow the synchronization of recorded and modeled behavior, the more likely it has happened. This can be further reinforced by defining this likelihood per object role, available from domain knowledge. For example, a certain object may be essential in the event, and its absence should not allow for partial matching.

The matching relation as defined in Def.~\ref{def:match} states that two elements $e \in \bar L_\varepsilon$ and $t_\mode \in \bar{\run}_\varepsilon$ are potential matches according to a matching function $\match$, \ie ${e \sim t_\mode} \iff \match(e,t_\mode) \geq \theta$ for some $\match$ and $0 \leq \theta \leq 1$. So far, $\match$ is defined on the similarity of the activity and transition label of $e$ and $t_\mode$ respectively, as well as the involved object roles in both. When talking about alignments, so far, $\theta$ is set to 1, meaning that all involved objects should match.
However, we can adapt this threshold to allow partial synchronization of recorded and modeled behavior by using the object role projections of $M$ discussed in Sec.~\ref{sec:model}. This means that we can adjust $\theta$ such that $0 \leq \theta \leq 1$ to compute how many of the involved objects can be synchronized.

\subsection{Relaxed alignment moves through projections}
Recall that a projection of $L$ onto objects $O \leq \Ob$ shows the recorded behavior involving only these objects. Similarly, the projection of $M$ onto objects $O$ models the behavior from the perspectives of only these objects, ignoring any restrictions for transition firings imposed by the modeled behaviors from other objects, \ie $\Ob - O$. Similarly, through alignments on these projections, we can compute how many of the involved objects can be synchronized. For example, if we look at system behavior $S_1$ from Fig.~\ref{fig:S1}. This is not behavior in the complete model $M$, since $\orderdepot{}$ and $\ring{}$ are not allowed to be present in the same run of $M$. However, if we look at $S_1$ from the perspective of just the deliverer $d_1$, it perfectly matches the modeled behavior from the corresponding model projection $M\proj_{d_1}$ of the deliverer.

\subsubsection{Relaxing the process model}
\begin{figure}[tb]
\centering
\begin{subfigure}{.6\textwidth}
  \centering
  \hspace*{-.5cm}
  \includegraphics[scale=0.6]{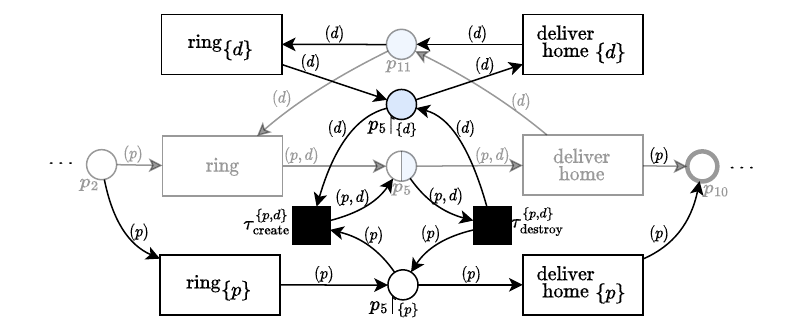}
  \hspace*{.5cm}
  \caption{Fragment of relaxed t-PNID process model $\tilde{M}$ of $M$ from Fig.~\ref{fig:re_net}.}
  \label{fig:projections_correlation}
\end{subfigure}
\begin{subfigure}{.99\textwidth}
  \centering
  \includegraphics[scale=0.7]{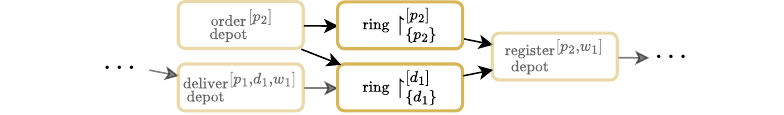}
  \caption{Fragment of $\tilde{L}_1$: a relaxed version of $L_1$ from Fig.~\ref{fig:L1}.}
  \label{fig:rel_log_fragment}
\end{subfigure}
\caption{Relaxations of a process model and a system log.}
\label{fig:relaxations}
\end{figure}

To allow for partial synchronization of events and transition firings, we introduce the concept of relaxing a process model. Relaxations add behavior through the imputation of projections into the respective t-PNID model. The relaxed process model $\tilde{M}=(\tilde{N}, m_i,m_f)$ extends a process model $M=(N,m_i,m_f)$ by directly adding its projections on all objects while integrating a \emph{correlation creation/destruction net} denoted as $N^C$, giving $\tilde{N} = (\bigcup_{O \subseteq \Ob} N\proj_O) \cup N^C$.
A fragment of $\tilde{M}$ of the running example is shown in Fig.~\ref{fig:projections_correlation}, for its formal definition and construction, we refer to~\cite{sommers2024conformance}. $\tilde{M}$ differs from $M$ by the projected transitions for $\ring{}$ and $\deliverhome{}$. As both transitions involve two objects, there are two projections for each. In $M$, there is a correlation in place $p_5$ between the package and the deliverer created after the firing of $\ring{}$ and is used in the firing of $\deliverhome{}$. The added silent transitions alternatively allow for this creation and destruction, enabling for example the behavior of $\seq{\ring{}\proj_{\set{d}}^{\mset{d_1}}, \ring{}\proj_{\set{p}}^{\mset{p_7}}, (\tau^{\set{p,d}}_{\text{create}})^{\mset{p_7,d_1}}, \deliverhome{\mset{p_7,d_1}}}$.

Note that without specific transitions for creating and destroying correlations between objects, certain transitions relying on such correlations may be undesirably disabled. Take for example the transition firings $\ring{\mset{p_1}}\proj_{p}$ and $\ring{\mset{d_1}}\proj_{d}$. Since $\deliverhome{}$ relies on the correlation from $p_5$, it is impossible to fire this transition with $p_1$ and $d_1$. Instead, the correlation creation/destruction net includes a silent transition $\tauc^{p_5}$ that can create the correlation between $p_1$ and $d_1$ after firings $\ring{\mset{p_1}}\proj_{p}$ and $\ring{\mset{d_1}}\proj_{d}$, enabling $\deliverhome{\mset{p_1,d_1}}$.

With each object projection added to $M$, there is the option to traverse the net for any combination of objects not relying on synchronization between them through their interaction. The additional silent transitions from $N^C$ to create and destroy correlation allow for moving between projections in the relaxed model. Note that $N^C$ only creates new silent transitions and arcs in $\tilde{N}$.

Aligning with a relaxed process model enables both relaxed model moves of the form $(\gg, t_\mode\proj_O)$, and relaxed synchronous moves of the form $(e, t_\mode\proj_O)$ following the same potential matching function as regular moves (\cf Def.~\ref{def:align_moves}), \ie $e = (a, O)$ and $O = \obj(t_\mode\proj_O)$. This enables us to precisely identify which objects require the execution of an activity in the model where the matching recorded event is missing and also shows that the observed and modeled behavior agree on the event, even if some involved objects are missing from the recorded event.

\subsubsection{Relaxing the system log}
Similarly, for the log, we allow for relaxations in terms of the synchronization of objects in recorded events. Recall from Sec.~\ref{sec:system_behavior} that the system log is a combination of traces that converge and diverge through their shared events. By removing these synchronization points through projections this can be relaxed.
Take for example $L_1$ from Fig.~\ref{fig:L1}. A fragment of a possible relaxed version $\tilde{L}_1$ is shown in Fig~\ref{fig:rel_log_fragment}, removing the synchronization in the recorded event of $\ring{\mset{p_2,d_1}}$ by replacing it with two concurrent projected events $\ring{}\proj_{\set{p_2}}^{\mset{p_2}}$ and $\ring{}\proj_{\set{d_1}}^{\mset{d_1}}$.
Note that many relaxed versions exist, as each event $e = (a, O) \in L$ can be relaxed for every partition of $O$.

We define $\tilde{L}$ to be a relaxed version of a system log $L$, if and only if it adheres to the following properties:
\begin{align}
    \forall_{o \in \Ob} ~& L_o = \tilde{L}\proj_o \label{eq:rellog_1} \\ 
    \forall_{(a, O) \in \bar{L}} ~& O = \sum_{(a, O)\proj_{O'} \in \bar{\tilde{L}}} O' \label{eq:rellog_2}
\end{align}
Eqs.~\ref{eq:rellog_1} and~\ref{eq:rellog_2} ensure respectively that the partial order is intact and that the relaxations correspond to actual partitions of the involved objects.

Relaxing the log enables both relaxed log moves of the form $(e\proj_O, \gg)$, and relaxed synchronous moves of the forms $(e\proj_O, t_\mode)$ and $(e\proj_O, t_\mode\proj_O)$ again with the same potential matching function used before. Similarly to the previously introduced relaxed moves, this enables us to precisely identify which objects cannot be mimicked by the process model for a recorded event, and that the observed and modeled behavior agree on the event, even if it involves only a subset of the objects for the model, or both the model and log.

\subsubsection{Relaxing the objects matching function}
Lastly, we add relaxations in the synchronous moves that allow for a mismatch in object names when a substitute object is available of the same role.
The \emph{substitute synchronous moves} emphasize a partial difference in the matching of the object names. An example is $\overline{(e, t_\mode)}^R$ with $e = (\ring{}, \mset{p_1, d_1})$, $\ell(t) = \ring{}$, $\obj(t_\mode) = \mset{p_1, d_2}$, and with $R = \set{d}$ to relax the matching of the deliverers. 

Aligning with a relaxed version of the matching function the object names enables synchronous moves of the form $\overline{(e, t_\mode)}^R$ and $\overline{(e\proj_O, t_\mode\proj_{O'})}^R$, completing the relaxations of moves. The potential matching function for substitute synchronous moves is slightly different since the objects of roles in $R$ should not match in the recorded event and transition firing. Instead, with $e = (a, O)$, it is defined as $a = \ell(t)$ and $(O\proj_{\R \setminus R} = \obj(t_\mode)\proj_{\R \setminus R})$ similar to the other synchronous moves, for roles not in $R$, and $(O\proj_R \cap \obj(t_\mode\proj_{O'})\proj_R = \emptyset) \wedge (|O\proj_R| = |\obj(t_\mode\proj_{O'})\proj_R|)$ for roles in $R$, \ie involved objects in $e\proj_O$ and $t_\mode\proj_{O'}$ are disjoint sets of the same size.

These relaxations for the model and the log are obtained from their projections and allow for partial matching regarding the behavior of involved objects. It is important to note that these relaxations only add behavior, as they only remove constraints on the synchronization of objects. A proof of this property is presented in~\cite{sommers2024conformance}.
\newline

To summarize, in this section, we introduced the notion of relaxed moves, enabled by projections on the model and log, defined formally as:
\begin{definition}{(Relaxed moves)}
Let $\tilde{L}$ be any relaxed version of a system log $L$ and $\tilde{M} = (\tilde{N}, m_i, m_f)$ a relaxed process model.
$\tilde{T}_\mode$ denotes the set of all possible transition firings in $\tilde{M}$~(see Def.~\ref{def:marking}).
$\tilde\al_s = \set{(e\proj_O, t_\mode\proj_{O'}) \mid e\proj_O = (a,O'')\proj_O \in \tilde{L}, t_\mode\proj_{O'} \in \tilde{T}_\mode, a = \ell(t), O = \obj(t_\mode\proj_{O'}), (O < O'' \vee O' < \obj(t_\mode))}$ is the set of possible \emph{relaxed synchronous moves}, following the potential matching function from Def.~\ref{def:match}, where at least one of $e\proj_O$ and $t_\mode\proj_{O'}$ is a projection.
$\tilde\al_l = \set{(e\proj_O, \gg) \mid e\proj_O = (a, O')\proj_O \in \tilde{L}, O < O'}$ is the set of possible \emph{relaxed log moves}, \ie moves on projection events.
$\tilde\al_m = \set{(\gg, t_\mode\proj_O) \mid t_\mode\proj_O \in \tilde{T}_\mode, O < \obj(t_\mode)}$ is the (infinite) set of possible \emph{relaxed model moves}, \ie firings of projection transitions in $\tilde{M}$.
$\bar\al_s^R = \set{(e\proj_O, t_\mode\proj_{O'}) \mid e\proj_O = (a, O'') \in \tilde{L}, t_\mode\proj_{O'} \in \tilde{T}_\mode, a = \ell(t), O\proj_{\R \setminus R} = \obj(t_\mode\proj_{O'})\proj_{\R \setminus R}, O\proj_R \cap \obj(t_\mode\proj_{O'})\proj_R = \emptyset, |O\proj_R| = |\obj(t_\mode\proj_{O'})\proj_R|}$ is the set of possible \emph{substitute synchronous moves}.
\end{definition}

\subsection{Relaxed alignment as a more accurate representation of reality}
When there is a discrepancy between recorded and modeled behaviors on a single perspective, regular deviating moves (log and model moves) are mandatory in the setting without relaxed moves. Contrarily, incorporating relaxed moves allows us to provide information on which perspectives the behaviors agree or disagree.
Further, it allows us to categorize the deviations based on the number of complying perspectives, which determines the likelihood of belonging to the system behavior. We denote this the \emph{relaxed alignment}, which can incorporate both the regular moves as well as the model and synchronous moves from projections and the substitute synchronous moves (relaxed moves):
\begin{definition}{(Relaxed alignment)}\label{def:rel_al}
Let $L=(\bar{L},\prec_L)$ be a system log and $\tilde{M} = (\tilde{N}, m_i, m_f)$ be a relaxed process model of $M = (N,m_i,m_f)$. A \emph{relaxed alignment} $\relal$ is a poset $\relal=(\bar\relal, \prec_{\relal})$, where $\bar\relal \subseteq (\al_l \cup \al_m \cup \al_s \cup \tilde\al_s \cup \tilde\al_l \cup \tilde\al_m \cup \bar\al_s^R)$, having the following properties:
\begin{enumerate}
    \item $\relal\proj_L$ is a relaxed version of $L$, \ie respecting Eqs.~\ref{eq:rellog_1} and~\ref{eq:rellog_2};\label{prop:relalign1}
    \item $m_i \xrightarrow{\relal\proj_T} m_f$, \ie $\forall_{\sigma \in (\al\proj_T)^<}, m_i \xrightarrow{\sigma} m_f$, so $\relal\proj_T$ is a run in the relaxed process model $\tilde{M}$ \label{prop:relalign2}
\end{enumerate}
with alignment projections on the log events $\relal\proj_L$ (\cf Eq.~\ref{eq:logrun}) and on the transition firings $\relal\proj_T$ (\cf Eq.~\ref{eq:modelrun}).
\end{definition}
The relaxed alignment allows for partial synchronization of the objects' individual behaviors through the included projections.
We aim to match as many perspectives as possible, assuming that compliance of representations indicates trust in the content, and therefore is more likely to be included in the system $S$. We accomplish the by imposing costs on the moves such that a move has lower cost when it synchronizes more objects. We define our cost function as follows and show that with this definition, the desired criteria are respected. For (relaxed) move $(e\proj_O, t_\mode\proj_O) \in \relal$ and some small $0 < \epsilon \ll 1$:
\begin{equation}\label{eq:rel_cost}
    c\stup{\stup{e\proj_O, t_\mode\proj_O}} =
    \begin{cases}
        \epsilon^2 & \text{if } t \in T^C \\
        |O| + (|\Var(t)| - |O|)\epsilon & \text{if } \gg \in \set{e\proj_O, t_\mode\proj_O} \wedge \ell(t) \neq \tau \\
        (|\Var(t)| - |O|)\epsilon & \text{otherwise}
    \end{cases}
\end{equation}
where $T^C$ denotes the transitions in the correlation creation/destruction net $N^C$ introduced in Sec.~\ref{sec:partial}. $c$ imposes different costs for three types of moves. From bottom to top, firstly, synchronous moves are penalized by the magnitude of the relaxation in terms of $\epsilon$, defined by the original number of involved objects of the transition in the process model minus the number of involved objects left in the projected transition, \ie $|\Var(t)| - |O|$. Secondly, the deviating moves are penalized by the number of involved objects ($|O|$) together with the relaxation magnitude in terms of $\epsilon$. Lastly, model moves on transitions from $N^C$ are of the lowest non-zero cost, set at $\epsilon^2$. We assume $\epsilon$ is small enough such that $x\cdot \epsilon^2 < \epsilon$ for any realistic $x \in \Nat$.

The cost function is specified as such that an optimal relaxed alignment synchronizes as much behavior as possible for as many objects simultaneously as possible. It does so through three criteria: (1) the number of distinguishable objects involved in deviating moves is minimized as a first-order criterion. (2) The alignment is minimally relaxed, \ie it synchronizes behavior between objects where possible, such that a projected move indicates that only the involved objects allow for the behavior, and (3) the correlations are kept intact as much as possible, so silent transitions from the correlation create/destroy net are only fired when it results in a decrease in the first two criteria. In~\cite{sommers2024conformance}, we prove that $c$ adheres to these criteria.

The optimal relaxed alignment, \ie the alignment maximizing and minimizing respectively the involved objects in synchronous moves and deviating moves (log and model moves), can be computed using the same methods as for computing regular alignments, as discussed in Sec.~\ref{sec:alignments}.

Going back to the list of issues discussed in Sec.~\ref{sec:issues}, we discuss how the optimal relaxed alignments result in more informative and accurate representations of the system $S$:
\begin{itemize}
    \item[\Lissue{mi}{e}] 
    $L_1$ and $\run_1$ from Fig.~\ref{fig:congruence1} both have a missing event with regard to the system behavior $S_1$, \ie $(\ring{\mset{p_1,d_1}}, \varepsilon_{S_1}) \in \cong_{L_1,S_1}$ and $(\ring{\mset{p_2,d_1}}, \varepsilon_{S_1}) \in \cong_{\run_1,S_1}$, which appear respectively as a model move and a synchronous move in the regular alignment $\al_{L_1,\run_1}$. With cost function $c$ from Eq.~\ref{eq:rel_cost}, the corresponding optimal relaxed alignment, as depicted in Fig.~\ref{fig:congruence1_relaxed}, resolves the recording issue identically to the regular alignment, inserting a regular model move (\ie including all originally involved objects), and resolves the modeling issue with a partial synchronous move, only including the deliverer object role. The regular model move $(\varepsilon_{L_1}, \ring{\mset{p_1,d_1}}) \in \relal_1$ (annotated yellow) shows that the activity fits the behavior for all objects in the model, even though it is not observed, suggesting that the activity could have been executed and there is a recording error. The partial synchronous move $(\ring{\mset{d_1}}_{\set{d}}, \ring{\mset{d_1}}_{\set{d}}) \in \relal_1$, partial log move $(\ring{\mset{p_2}}_{\set{p}},\varepsilon_{\tilde{\run}_1}) \in \relal_1$ and partial model moves $(\varepsilon_{L_1}, \tautwo{\mset{d_1}}_{\set{d}}), (\varepsilon_{L_1}, \tauone{\mset{p_1}}_{\set{p}}) \in \relal_1$ (annotated purple) shows that the recorded behavior fits the behavior for a subset of the involved objects in the corresponding activity. Note that this relaxed alignment aligns the relaxed process model $\tilde{M}$ (\cf Fig.~\ref{fig:projections_correlation}) with the relaxed version $\tilde{L}_1$ of $L_1$ as shown in Fig.~\ref{fig:rel_log_fragment}. 
    $\tilde{\run}_1$ allows the package $p_2$ to use $\tauone{}$ and the deliverer $d_1$ to use $\ring{}$ and $\tautwo{}$, after which the correlation is created in $\substack{\tau_{\text{create}}^{\set{p,d}}}^{\mset{p_2,d_1}}$ and the synchronized behavior can continue in $\deliverdepot{\mset{p_2,d_1,w_1}}$.
    \begin{figure}[tb]
        \centering
        \hspace*{-1.7cm} 
        \includegraphics[scale=0.6]{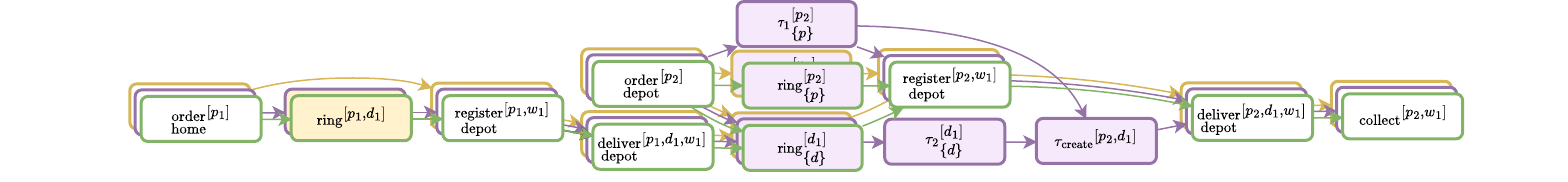}
        \hspace*{1.7cm} 
        \caption{Optimal relaxed alignment $\relal_1 = \relal_{L_1,M}$ between $L_1$ and $M$ with regard to $S_1$, \ie $\cong_{\relL_1,\relrun_1,S_1} \iff \cong_{\relal_1,S_1}$.}
        \label{fig:congruence1_relaxed}
    \end{figure}
    \item[\Lissue{in}{e}]
    The incorrectly recorded event $\ring{\mset{p_3,d_1}}$ appeared as a log move in the regular alignment (see Fig.~\ref{fig:congruence2}) as well as in the relaxed alignment, shown in Fig.~\ref{fig:congruence2_relaxed}, \ie $(\ring{\mset{p_3,d_1}}, \varepsilon_{\tilde{\run}_2}) \in \relal_2$ (annotated yellow). It is a log move because none of the involved objects allow the behavior, and the event can not even be partially synchronized to the model. The $\ring{}$ activity is necessary for package $p_4$ according to the model, resulting in model moves for both $\ring{}$ and $\registerdepot{}$ and a $\registerdepot{}$ log move in the regular alignment. Since the relaxed alignment allows model moves on the projection net, the firing of $\ring{\mset{p_4}}_{\set{p}}$ and $\ring{\mset{d_1}}_{\set{d}}$ independently allows for a synchronous move on $\registerdepot{}$ (annotated purple).
    \begin{figure}[tb]
        \centering
        \hspace*{-2.2cm} 
        \includegraphics[scale=0.6]{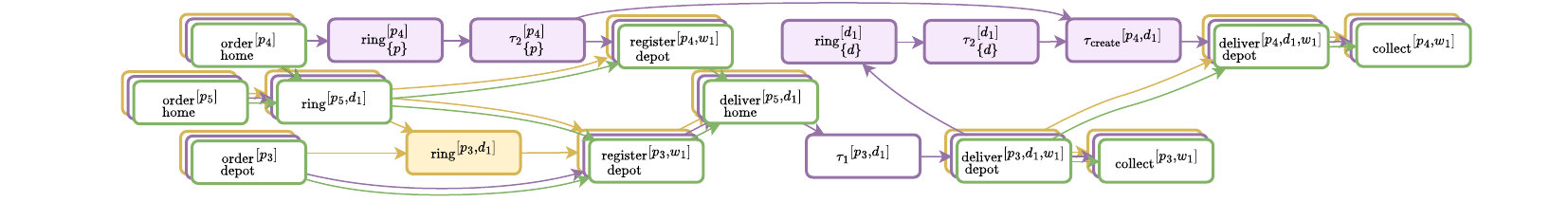}
        \hspace*{2.2cm} 
        \caption{Optimal relaxed alignment $\relal_2 = \relal_{L_2,M}$ between $L_2$ and $M$ with regard to $S_2$, \ie $\cong_{\relL_2,\relrun_2,S_2} \iff \cong_{\relal_2,S_2}$.}
        \label{fig:congruence2_relaxed}
    \end{figure}
    \item[\Lissue{mi}{o}]
    By definition, there is no atomic synchronous move possible corresponding to the recorded event with the missing deliverer object $\deliverdepot{\mset{p_6,w_2}} \in L_3$, resulting in a pair of log and model move $(\deliverdepot{\mset{p_6,w_2}}, \varepsilon_{\run_3}), (\varepsilon_{L_3}, \deliverdepot{\mset{p_6,d_2,w_2}}) \in \al$ (see Fig.~\ref{fig:congruence3}). Since the relaxed alignment allows for partial matching on projections for certain combinations of object roles, this issue is resolved by a partial synchronous move for the package and warehouse $(\deliverdepot{\mset{p_6,w_2}}_{\set{p,w}}) \in \relal_{L_3,\run_3}$ and a model move for the deliverer in parallel $(\varepsilon_{L_3}, \deliverdepot{\mset{d_2}}_{\set{d}}) \in \relal_{L_3,\run_3}$ (see Fig.~\ref{fig:congruence3_relaxed}).
    \begin{figure}[tb]
        \centering
        \hspace*{-0.3cm} 
        \includegraphics[scale=0.6]{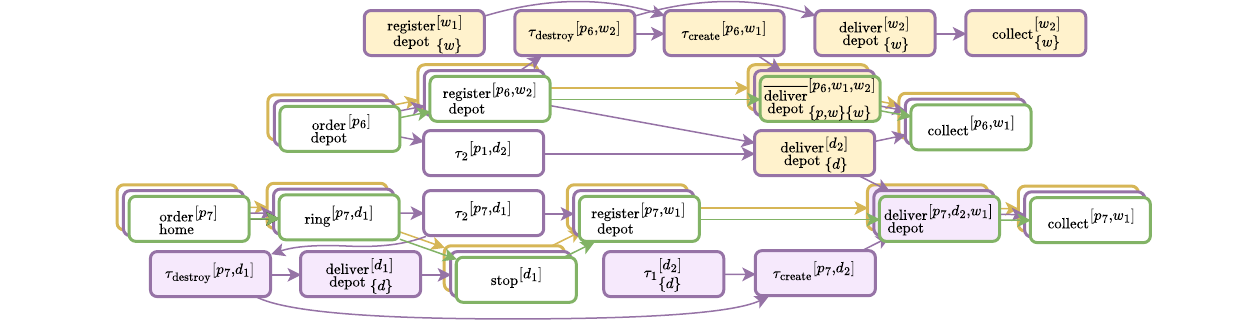}
        \hspace*{0.3cm} 
        \caption{Optimal relaxed alignment $\relal_3 = \relal_{L_3,M}$ between $L_3$ and $M$ with regard to $S_3$, \ie $\cong_{\relL_2,\relrun_3,S_3} \iff \cong_{\relal_3,S_3}$.}
        \label{fig:congruence3_relaxed}
    \end{figure}
    \item[\Lissue{in}{o}]
    The issue regarding the incorrect association of involved objects is similarly resolved as the missing objects, where a synchronous move is not possible because the objects do not match, resulting again in a log and model move pair. As discussed before, with this construct it is unclear which object was involved in the actual execution. For package $p_6$, the deliverer logged a different warehouse for its delivery, as depicted by the $\deliverdepot{}$ activities in $L_3$, $S_3$, and $\run_3$ in Fig.~\ref{fig:congruence3}. This issue is exposed in the relaxed alignment with a move on the substitute synchronous move on the warehouse $\overline{\deliverdepot{}}_{\set{p,w}\set{w}}^{\mset{p_1,w_1,w_2}}$, as shown in Fig.~\ref{fig:congruence3_relaxed}. Note that for this move to be incorporated, relaxed model moves are required for the warehouse object role (annotated yellow). The correlation between package $p_6$ and warehouse $w_2$ is switched to $w_1$ by the silent moves $\substack{\tau_{\text{destroy}}^{\set{p,w}}}^{\mset{p_6,w_2}}$ and $\substack{\tau_{\text{create}}^{\set{p,w}}}^{\mset{p_6,w_1}}$. Where the handover of package $p_7$ from deliverer $d_1$ to $d_2$ results in $\tstop{}$ activity to be delayed for $d_1$ in the regular alignment (see Fig.~\ref{fig:congruence3}), the relaxed alignment allows for each recorded event to be synchronized to the model by the correlation switch $\substack{\tau_{\text{destroy}}^{\set{p,d}}}^{\mset{p_7,d_1}}$ and $\substack{\tau_{\text{create}}^{\set{p,d}}}^{\mset{p_7,d_2}}$ with additional relaxed model moves to move the deliverers accordingly.
    \item[\Lissue{mi}{p}]
    This issue is resolved in relaxed alignment similar to regular alignments as discussed in Sec.~\ref{sec:issues}.
    \item[\Lissue{in}{p}]
    The recorded event $\ring{\mset{p_8,d_1}} \in L_4$ shown in Fig.~\ref{fig:congruence4} is logged at a different position in time than where it actually happened according to $S_4$. Since none of the involved object roles allow this ordering of activities, the relaxed alignment, depicted in Fig.~\ref{fig:congruence4_relaxed} shows the same model and log move pair as the regular alignment shown in Fig.~\ref{fig:congruence4}. On the other hand, the multitasking behavior of deliverer $d_1$ for packages $p_9$ and $p_{10}$ only deviates from the modeled behavior for the deliverer object role. Therefore, the relaxed alignment $\relal_4$ allows for a partial synchronous move on $\ring{\mset{p_9}}_{\set{p}}$ with corresponding partial log move $\ring{\mset{d_1}}_{\set{d}}$. To synchronize the behavior at the $\deliverhome{}$ activity again, $\relal_4$ contains a partial model move $(\varepsilon_{L_4}, \ring{\mset{d_1}}_{\set{d}})$ and a create correlation move $\substack{\tau_{\text{create}}^{\set{p,d}}}^{\mset{p_9,d_1}}$.
    \begin{figure}[tb]
        \centering
        \hspace*{-1.2cm} 
        \includegraphics[scale=0.6]{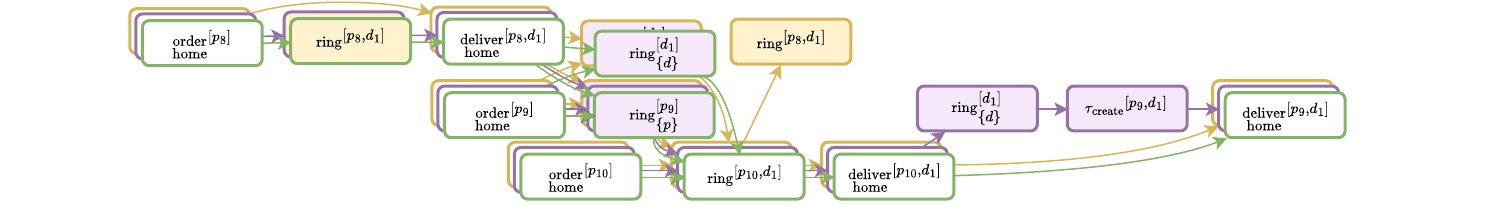}
        \hspace*{1.2cm} 
        \caption{Optimal relaxed alignment $\relal_4 = \relal_{L_4,M}$ between $L_4$ and $M$ with regard to $S_4$, \ie $\cong_{\relL_4,\relrun_4,S_4} \iff \cong_{\relal_4,S_4}$.}
        \label{fig:congruence4_relaxed}
    \end{figure}
\end{itemize}

The relaxed moves, \ie model and synchronous moves on projections and synchronous moves with substituted object names, allow for more interpretation in the alignment of recorded and modeled behavior. The relaxed alignment results in a closer match between log and model, as it maximizes the information between the two representations, and potentially leads potentially to a more accurate representation of the underlying process.

\subsection{Trusting representations locally}
A relaxed alignment can uncover issues in both recorded and modeled behavior, and it does so locally. This means that one does not need to make a choice and fully trust either the log or the model, but rather one can determine trust levels locally, based on the specific issues found in the data. 
After identifying issues in certain parts of the process with respect to certain perspectives, it is possible to generate statistics in order to establish trust levels for those specific contents in both the log and the model. Information on which contents is of low trust and which content is of high trust is valuable for log and model repair~\cite{fahland2015model,rogge2014temporal,rogge2013improving,wang2015cleaning}, as the repair can get a clear focus on the problematic content.

Furthermore, the relaxed alignment $\relal$ gives the basis for the analysis of what behavior is most likely to have happened in reality, through the classification of the (relaxed) moves such as the last row of Tab.~\ref{tab:trusts}. This improved representation of $S$ is valuable for decision point analysis and performance analysis, since ideally, only the behavior of $S$ should be analyzed.

\section{Conclusion}\label{sec:conclusion}
Real-life processes involve multiple objects affecting each other's behaviors through interactions in shared activities. While the exhibited behavior of the process is an unknown, invisible entity, we described how it is represented by its recorded behavior (a system log) and modeled behavior (a normative process model), both coming with potential quality issues.

Alignments can provide a more trustworthy and richer representation of the process, by reconciling the recorded and modeled behavior. We have shown that alignments of a system log and a model are capable of exposing deviations in inter-object dependencies, which already provides a more accurate description of reality as opposed to the individually aligned objects. 

With possible quality issues in both the recorded and modeled behavior, interpretations of deviations can be ambiguous. Through relaxations, specifically on interacting perspectives, we enable partial matching of log and model, to more precisely pinpoint potential sources of deviations. Depending on the number of perspectives the behaviors agree on, we classify the deviations. This allows to identify the trustworthy and untrustworthy contents of the two representations, achieve a better understanding of the underlying process, and expose quality issues.

In this work, we have presented the relaxations in a general sense, where constraints from each of the perspectives can be relaxed. In a real-life setting, it is interesting to take into account how realistic these relaxations are based on domain knowledge, which can be integrated into the cost function for relaxed moves. For future work, we plan to evaluate the added value of relaxed system alignments in the settings of performance analysis and decision point analysis.

\textbf{Acknowledgements.}
This work is done within the project ``Certification of production process quality through Artificial Intelligence (CERTIF-AI)'', funded by NWO (project number: \href{https://www.nwo.nl/projecten/17998}{17998}).

\bibliographystyle{plain}
\bibliography{main}

\end{document}